\documentclass[conference,a4paper]{IEEEtran}
\ifCLASSOPTIONcompsoc
\usepackage[nocompress]{cite}
\else
\usepackage{cite}
\fi
\ifCLASSINFOpdf
\else
\fi
 
\usepackage{times}
\usepackage{amsmath}
\usepackage{amssymb}
\usepackage{array}
\usepackage{url}
\usepackage{comment}
\usepackage{rotating}
\usepackage{multirow}
\usepackage{graphicx}
\usepackage{subfigure}
\usepackage{color}
\usepackage{acronym}
\usepackage{algorithmic}
\usepackage[normalem]{ulem}
\usepackage{epstopdf}
\usepackage[pagebackref=true,breaklinks=true,letterpaper=true,colorlinks,bookmarks=false]{hyperref}

\begin{document}

\title{Face Destylization}

\author{\IEEEauthorblockN{Fatemeh~Shiri}
\IEEEauthorblockA{Australian National University\\
fatemeh.shiri@anu.edu.au}
\and
\IEEEauthorblockN{Xin Yu}
\IEEEauthorblockA{Australian National University\\
xin.yu@anu.edu.au}
\and
\IEEEauthorblockN{Piotr~Koniusz}
\IEEEauthorblockA{Data61/CSIRO\\
piotr.koniusz@dada61.csiro.au}
\and
\IEEEauthorblockN{Fatih Porikli}
\IEEEauthorblockA{Australian National University\\
fatih.porikli@anu.edu.au}}
\maketitle


\begin{abstract}
Numerous style transfer methods which produce artistic styles of portraits have been proposed to date. However, the inverse problem of converting the stylized portraits back into realistic faces is yet to be investigated thoroughly. Reverting an artistic portrait to its original photo-realistic face image has potential to facilitate human perception and identity analysis. In this paper, we propose a novel Face Destylization Neural Network (FDNN) to restore the latent photo-realistic faces from the stylized ones. We develop a Style Removal Network composed of convolutional, fully-connected and deconvolutional layers. The convolutional layers are designed to extract facial components from stylized face images. Consecutively, the fully-connected layer transfers the extracted feature maps of stylized images into the corresponding feature maps of real faces and the deconvolutional layers generate real faces from the transferred feature maps. To enforce the destylized faces to be similar to authentic face images, we employ a discriminative network, which consists of convolutional and fully connected layers. We demonstrate the effectiveness of our network by conducting experiments on an extensive set of synthetic images. Furthermore, we illustrate our network can recover faces from stylized portraits and real paintings for which the stylized data was unavailable during the training phase. 
\end{abstract}

\IEEEdisplaynontitleabstractindextext
\IEEEpeerreviewmaketitle
\ifCLASSOPTIONcompsoc
\IEEEraisesectionheading{\section{Introduction}\label{sec:introduction}}
\else

\renewcommand{\thefootnote}{\fnsymbol{footnote}}
\footnotetext[1]{\label{foot:maps}This work has been published in DICTA'17.\vspace{-0.6cm}}

\section{Introduction}
\label{sec:introduction}
Applying artistic styles to existing photographs has attracted much attention in both academia and industry with several interesting applications. The inverse problem of reverting an artistic portrait back to its photo-realistic version is investigated in this paper. Revealing the latent real faces can provide essential information for human perception, computer analysis and photo-realistic multimedia content editing. Since facial details and expressions in stylized portraits often undergo severe distortions and become contaminated with artifacts such as profile edges and color changes e.g., as in Fig. \ref{fig:opena} and Fig. \ref{fig:opene}, recovering a photo-realistic face image from its stylized version is very challenging.  

The seminal work of \cite{gatys2016image} stylizes the content of an arbitrary image according to a given reference artwork and achieves appealing style transfer results, hovewer, its iterative optimization procedure is computationally costly. Several methods based on feed-forward neural networks \cite{ulyanov2016texture,ulyanov2016instance,johnson2016perceptual,dumoulin2016learned,li2017diversified,chen2016fast,zhang2017multi,huang2017arbitrary} accelerate the style transfer for specific styles. 

For our inverse problem, the above style transfer methods fail to recover authentic face images as shown in Fig.~\ref{fig:openf} and Fig.~\ref{fig:openg}. These approaches typically use Gram matrices to capture style-related contents. Since Gram matrices are designed to measure the correlations between feature maps of a style image and a target face, the spatial structure of an output image is not guaranteed to be similar to the target face. Therefore, existing style transfer methods which rely on Gram matrices are not sufficient for restoring photo-realistic portraits.


\begin{figure}[t]
\begin{minipage}{0.8\linewidth}
\centering
\subfigure[Seen input]{\label{fig:opena}\scalebox{1}[1]{\includegraphics[width=0.23\linewidth]{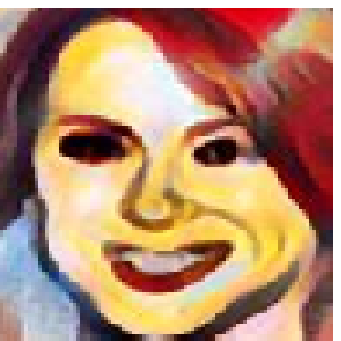}}}
\subfigure[Gatys\cite{gatys2016image}]{\label{fig:openb}\scalebox{1}[1]{\includegraphics[width=0.23\linewidth]{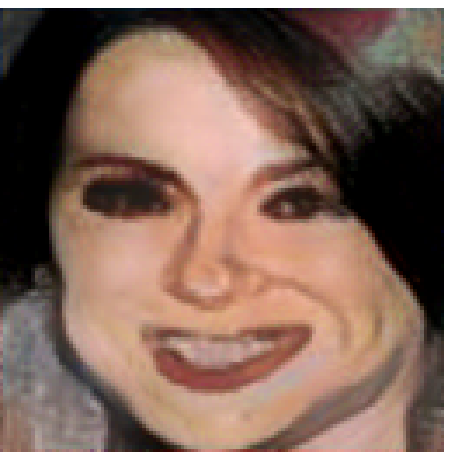}}}
\subfigure[Using \cite{johnson2016perceptual}]{\label{fig:openc}\scalebox{1}[1]{\includegraphics[width=0.23\linewidth]{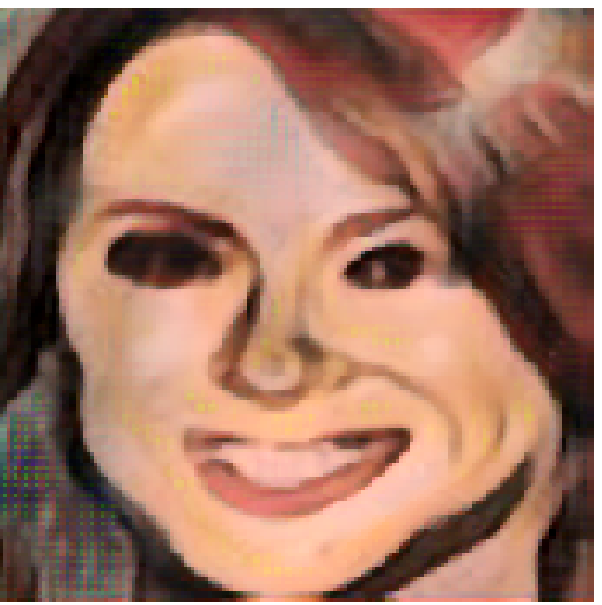}}}
\subfigure[Our result]{\label{fig:opend}\scalebox{1}[1]{\includegraphics[width=0.23\linewidth]{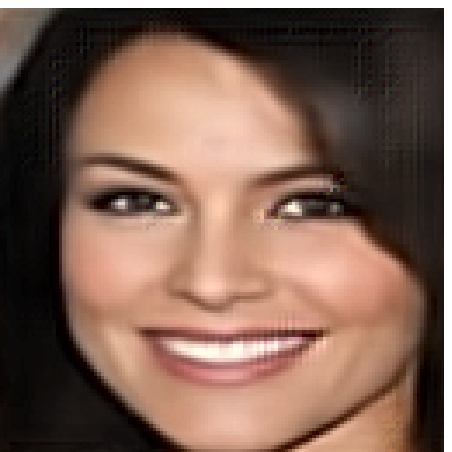}}}
\subfigure[Unseen in.]{\label{fig:opene}\scalebox{1}[1]{\includegraphics[width=0.23\linewidth]{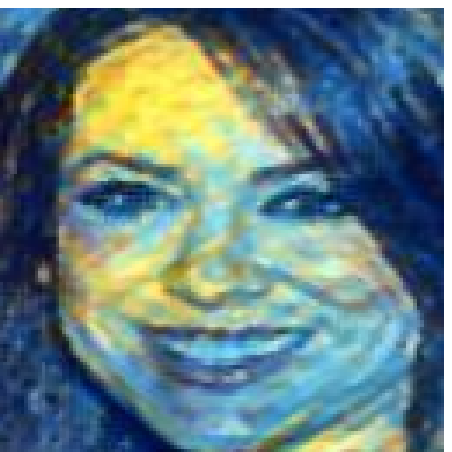}}}
\subfigure[Gatys\cite{gatys2016image}]{\label{fig:openf}\scalebox{1}[1]{\includegraphics[width=0.23\linewidth]{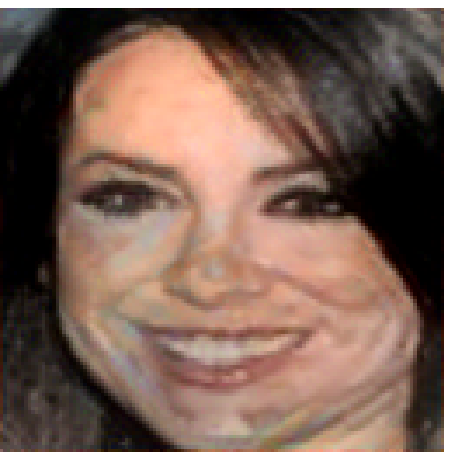}}}
\subfigure[Using \cite{johnson2016perceptual}]{\label{fig:openg}\scalebox{1}[1]{\includegraphics[width=0.23\linewidth]{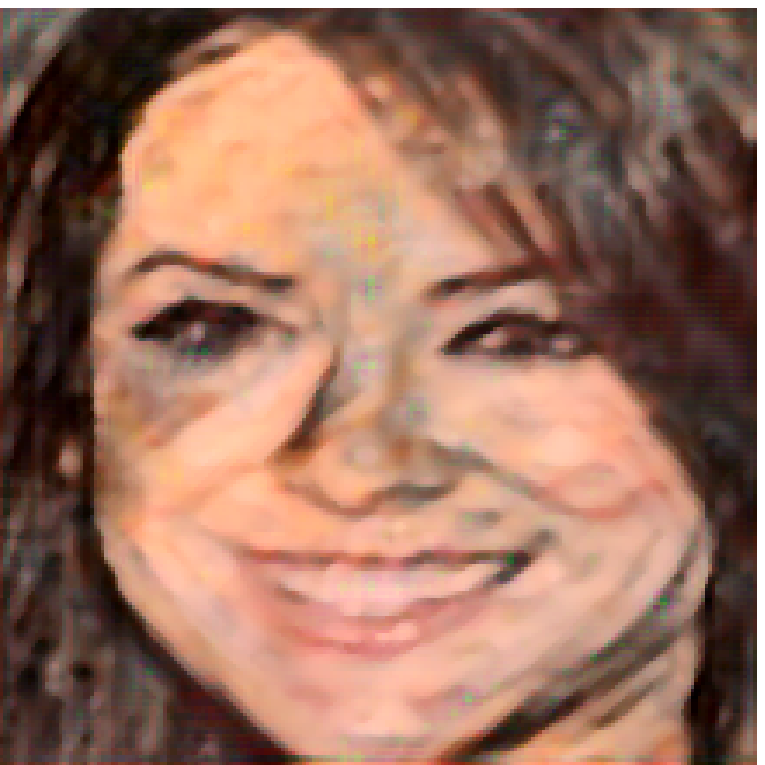}}}
\subfigure[Our result]{\label{fig:openh}\scalebox{1}[1]{\includegraphics[width=0.23\linewidth]{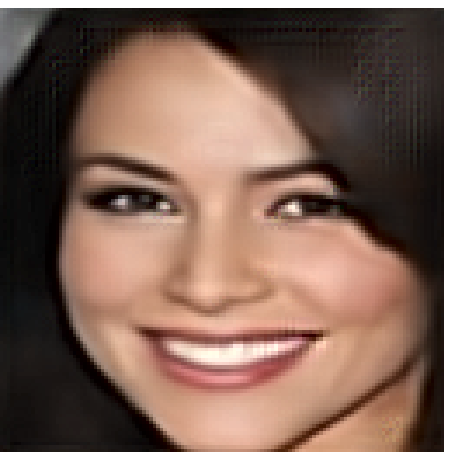}}}
\end{minipage}
\begin{minipage}{0.19\linewidth}
\centering
\subfigure[Original]{\label{fig:openi}\scalebox{1}[1]{\includegraphics[width=0.98\linewidth]{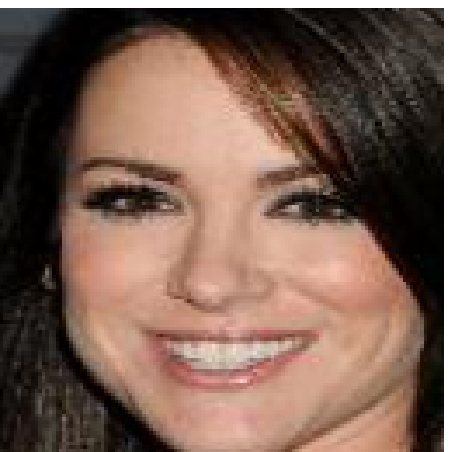}}}
\end{minipage}
\caption{Comparison to the state-of-art methods. (a) and (e) $128\times 128$ stylized face images in \emph{Candy} style (which is seen and used for training) and in \emph{Starry Night} style (which is unseen style), respectively. (b, f) Results obtained by applying \cite{gatys2016image} for the given stylized faces. (c, g) Results obtained by applying \cite{johnson2016perceptual}. (d, h) Our destylization results. (i) $128\times128$ ground-truth face image (used for evaluation purposes; not available to the algorithm for training).} 
\label{fig:open}
\end{figure}


To capture local statistics of a style image, some approaches use a so-called patch-based Generative Adversarial Network (GAN) \cite{li2016precomputed,isola2016image}. However, patch-based GANs do not take the global structure of faces into account thus a direct application of patch-GAN may not produce satisfactory results. We will show later that patch-based methods \cite{li2016precomputed, isola2016image} fail to attain the consistency of face colors. For the inverse problem, the patch-based GAN methods result in even bigger inconsistencies.


We note that the state-of-the-art style transfer methods~\cite{li2016precomputed,ulyanov2016texture,johnson2016perceptual} do not fully take into consideration how to extract facial features from different stylized images and then recover realistic face images.
Our goal is to reveal the latent real face images from multiple style portraits (seen styles) and achieve destylization even when the styles are not available in the training dataset (unseen styles). 



To this end, we propose a novel destylization network that automatically maps the stylized faces to photo-realistic ones in an end-to-end fashion. Our network is composed of two components: a generative part, named \emph{Style Removal Network (SRN)}, and a discriminative part. SRN constitutes convolutional, fully-connected and deconvolutional layers. The convolutional layers are exploited to extract facial components from stylized face images. As we aim to generate realistic face images, a fully-connected layer is developed to map the extracted feature maps of stylized faces to the feature maps of real faces. Then the mapped feature maps are projected to the image domain, thus forming face images. The discriminative network enforces the generated face images to lie in the same latent space as the realistic face images, in the manner similar to~\cite{Goodfellow2014, denton2015deep, yu2016ultra}. We train the entire network on a large-scale dataset of stylized and real face pairs. Our proposed framework can restore important facial details and attributes thanks to the style removal and discriminative subnetworks.


Furthermore, we observe that the filters of Convolutional Neural Network (CNN) learned during training (seen styles) are able to extract features from images containing unseen styles. Thus, the facial information of stylized portraits can be extracted and used to represent features of real faces. Therefore, our network can also restore the images of faces given an unseen style.
In the experimental section, we demonstrate that our network is able to recover realistic faces from both seen and unseen styles e.g., synthesized and original portraits and paintings.

Below, we summarize our main contributions:
\begin{itemize}
\item We propose FDNN which is able to generate photo-realistic faces from stylized ones. The results resemble accurately the ground-truth faces in terms of facial properties e.g., facial profiles and expressions. 
%
\item We develop a style removal sub-network to extract features from stylized input face images, then map these style features to real facial features and re-project them to the image domain for the purpose of generating authentic looking faces. 
\item We provide a dataset of pairs of the stylized and real face images used in our experiments to stimulate further research in destylization.
\end{itemize}

To the best of our knowledge, our framework is the first attempt to provide a unified approach for face destylization which can remove both seen and unseen styles (observed cf. unobserved styles during training). 


\section{Related Work}
\label{sec:Related Work}
Next, we briefly review deep generative image models, deep style transfer methods, and image translation approaches.

\subsection{Deep Generative Image Models}
Recently, several frameworks have been proposed for image generation, such as variational auto-encoders \cite{kingma2013auto}, auto-regressive models \cite{oord2016pixel}, and GANs \cite{Goodfellow2014}. Among these models, GANs generate impressive results because they employ adversarial losses that force the generated images to be indistinguishable from their real counterparts. In order to improve the stability of the training procedure of GANs, various methods have been proposed \cite{huang2016stacked,denton2015deep,isola2016image,reed2016generative,salimans2016improved,arjovsky2017wasserstein}.
GANs are also employed by the style transfer \cite{li2016precomputed} and cross-domain image generation \cite{bousmalis2016unsupervised,ioffe2015batch,liu2016coupled,liu2017unsupervised,kim2017learning} approaches. 
Li and Wand \cite{li2016precomputed} train a Markovian GAN for image style transfer such that a discriminative training is applied on Markovian neural patches to capture local style statistics.
However, patch-based methods may fail to capture the global structure of objects.

\subsection{Deep Style Transfer}
Style transfer methods transfer the style of a specific artwork into a given photograph. They can be divided into two categories: \emph{image optimization-based} and \emph{feed forward} methods.

The optimization-based method \cite{gatys2016image} transfers the style by updating pixels of the image iteratively. It minimizes the distance between Gram matrices generated from feature maps of the style and synthesized image with respect to input noise. Gram matrices capture so-called feature co-occurrences and they are popular in image recognition \cite{pk_tensor,pk_da,sparse_tensor_cvpr}. 
The approach \cite{yin2016content} initializes the optimization algorithm with a content image instead of noise. 
Li and Wand \cite{li2016combining} use Markov Random Field (MRF) in the deep feature space to enforce local patterns. The work \cite{gatys2016preserving} employs linear models to transfer styles and to preserve colors by matching color histograms. Gatys ~\emph{et al.}~\cite{gatys2016controlling} 
detect and control spatial, color and scale factors during the stylization process.
In \cite{wilmot2017stable}, the loss function is improved by imposing a histogram-based loss. The above optimization-based methods require a time-consuming iterative optimization process, which limits their practical application. 


In contrast, \emph{feed-forward} approaches replace the original on-line iterative optimization procedure by off-line training to produce stylized images through a single forward pass \cite{ulyanov2016texture,johnson2016perceptual,li2016precomputed}. Johnson \emph{et al.} \cite{johnson2016perceptual} train the generative network by perceptual loss functions. 
The architecture of their generator network follows work \cite{radford2015unsupervised}. However, they additionally use residual blocks and replace pooling layers by so-called fractionally strided convolutions.
In a concurrent work, \cite{ulyanov2016texture} use a multi-resolution architecture for their generator network.
Li and Wand\cite{li2016precomputed} pre-compute a Markovian GAN which captures the feature statistics of  
patches. To achieve faster convergence, Ulyanov \emph{et al.} \cite{ulyanov2016instance,ulyanov2017improved} replace batch with instance normalization in the generator.
These feed-forward approaches \cite{ulyanov2016texture,johnson2016perceptual,li2016precomputed,ulyanov2016instance} 
are three orders of magnitude faster than optimization-based style transfer methods. However, these networks only transfer images for a predefined style and they need to be re-trained for each new style. Some recent approaches improve the style transfer from a single style to multiple styles \cite{chen2016fast,dumoulin2016learned}. Dumoulin \emph{et al.} \cite{dumoulin2016learned} propose to train a style transfer network for multiple styles by the use of a conditional instance normalization. Given feature activations of the content and style images, \cite{chen2016fast} replaces the content features with the closest-matching style features patch-by-patch.
A recent summary of state-of-the-art stylization methods can be found in \cite{jing2017neural}.

\subsection{Image Transformation}
Mapping images from one domain to another has a wide range of applications. The idea of image transformation comes from so-called image analogies \cite{hertzmann2001image} which focus on the non-parametric patch-based texture synthesis from a single input-output training image pair. 
Methods \cite{isola2016image,yu2016ultra,sangkloy2016scribbler,karacan2016learning,denton2015deep,radford2015unsupervised,salimans2016improved} employ neural networks to learn a parametric translating function from a large dataset of input-output pairs, such as super-resolution and colorization. 
Isola \emph{et al.} \cite{isola2016image} propose the “pix2pix” framework to learn a mapping from input to output by a conditional GAN. Similar ideas have been applied to generating photographs from sketches \cite{sangkloy2016scribbler}, semantic layout and scene attributes\cite{karacan2016learning}.

Moreover, \cite{isola2016image} also uses a convolutional patchGAN classifier for its discriminator network. 
The above patch-based method does not take the global structure of faces into account. 
Furthermore, their network employs the architecture "Unet" to transfer the source to the target domain and utilizes low-level features in the generative part that can result in distorted facial images.
In contrast, our approach takes into account the global structure of faces and learns how to extract usuful features for face destylization.
 

\begin{figure*}[!t]
\centering
\scalebox{1}[1]{\includegraphics[width=1.0\linewidth]{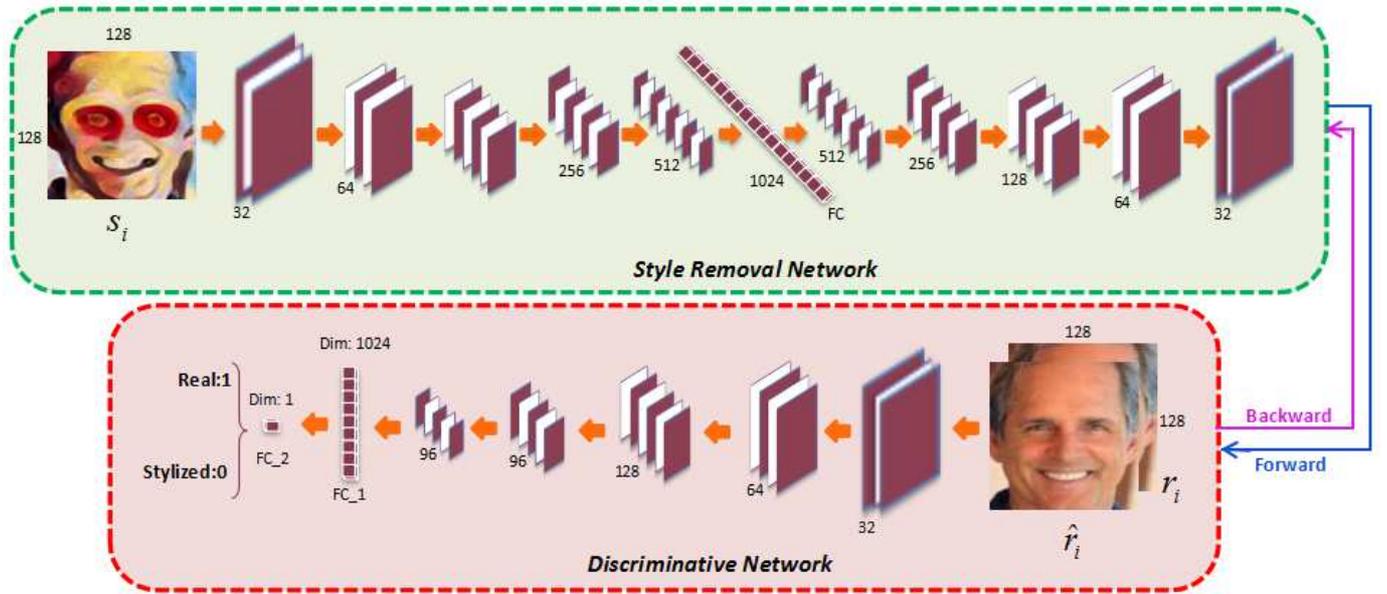}}
\caption{Face destylization neural network consists of two parts: a generative network (green frame) and a discriminative network (red frame).}
\label{fig:pipeline}
\end{figure*}
 
\section{Method}
Our FDNN network has two components: (i) a Style Removal Network (SRN), which transforms stylized faces to the photo-realistic ones, and (ii) a discriminative network, which enforces the generated faces by SRN to be indistinguishable from the real faces.
Figure~\ref{fig:pipeline} illustrates the overall architecture of our proposed network.  

\subsection{Style Removal Network}
In Fig.~\ref{fig:pipeline}, our SRN is enclosed by the green frame.
SRN aims at removing various styles of portraits and generating realistic faces. Our SRN comprizes convolutional layers followed by batch normalization layers, a fully connected layer and deconvolutional layers followed by batch normalization layers. 
The convolutional layers are employed to extract facial features from stylized face images. Then, we incorporate a fully-connected layer to transfer the extracted feature maps of stylized images into the feature maps of real faces. In order to synthesize images of real faces, deconvolutional layers project these transferred feature maps to the image domain.

In order train SRN, we use stylized portraits as inputs and their corresponding ground-truth images of real faces as desired supervising output signals. Since a dataset of portrait/real face pairs is not readily available, we opt to generate a large number of stylized faces in numerous styles from real face images. Figure ~\ref{fig:Desc} and Fig.~\ref{fig:Desf} illustrate the effectiveness of SRN. 

\subsection{Discriminative Network}
Using only Euclidean distance, i.e. $\ell_2$ loss, between the destylized faces and the corresponding ground-truth real ones tends to generate over-smoothed results as shown in Fig.~\ref{fig:Desc} and Fig.~\ref{fig:Desf}, and this phenomenon is also mentioned in~\cite{yu2016ultra}. Therefore, a class-specific discriminative objective is also incorporated into our SRN, aiming to enforce the destylized face images to lie on the same latent space of the authentic face images.

As shown in the red frame of Fig.~\ref{fig:pipeline}, the discriminative network is constructed by convolutional and fully connected layers. Its role is to determine whether an image is sampled from real face images or the destylized ones. With the help of the so-called discriminative adversarial loss, we can force generated destylized faces to be more similar to real ones. This is achieved by back-propagating the adversarial loss to update the parameters of SRN. Figure~\ref{fig:Desd} and Fig.~\ref{fig:Desg} illustrate the impact of the adversarial loss on the final results.

\subsection{Training Details}
\label{sec:training}

Our FDNN is trained in an end-to-end manner. We use Stylized Face (SF) and Real Face (RF) ground-truth image pairs  $(s_i,r_i)$
as our training dataset, where $r_i$ represents the real face images aligned by eyes only, and $s_i$ is a synthesized SF image from $r_i$. For each real face $r_i$, we generate eight different SFs i.e., Edvard Munch's \emph{Scream}, \emph{Candy}, \emph{Feathers}, \emph{Starry Night} by Van Gogh, \emph{la Muse} by Pablo Picasso, Wassily Kandinsky's \emph{Composition VII}, \emph{Mosaic} and Francis Picabia's \emph{Udnie}, and obtain SF/RF training pairs. The stylized faces of \emph{Scream}, \emph{Candy} and \emph{Feathers} are used in the training stage. 
As detailed in Sec.~\ref{Sec:Data}, we find that these distinct portraits provide a sufficient training data for our needs.

\begin{figure*}[t]
\centering
\includegraphics[width=0.10\linewidth]{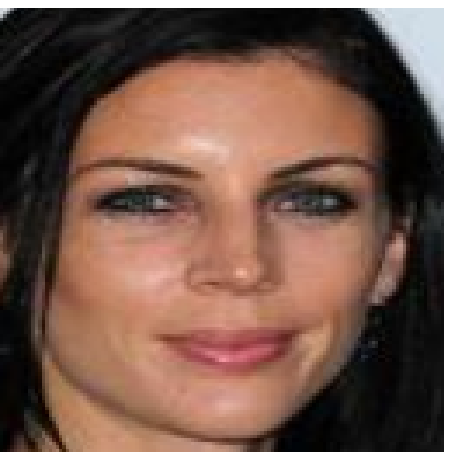}
\includegraphics[width=0.10\linewidth]{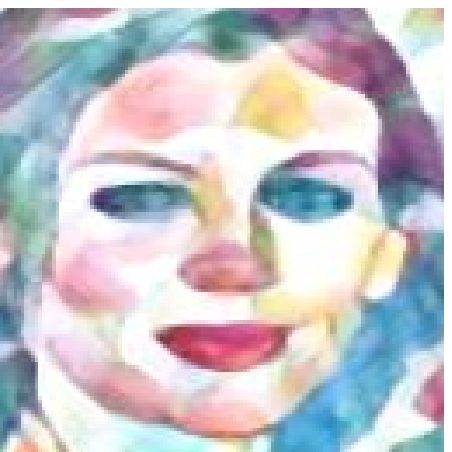}
\includegraphics[width=0.10\linewidth]{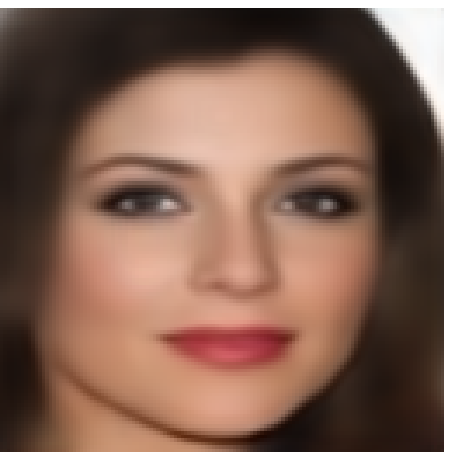}
\includegraphics[width=0.10\linewidth]{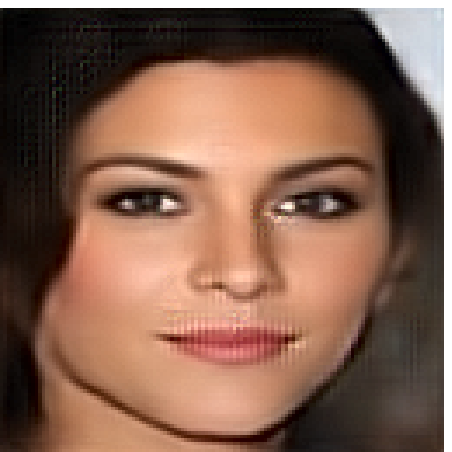}
\includegraphics[width=0.10\linewidth]{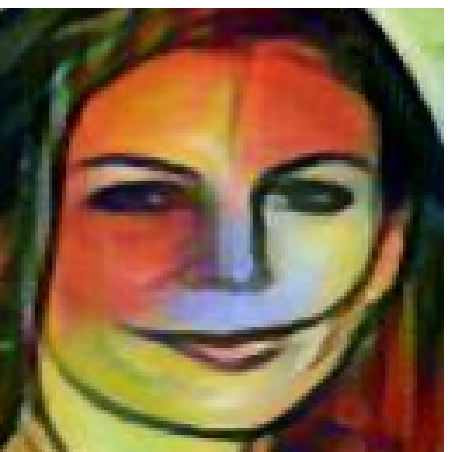}
\includegraphics[width=0.10\linewidth]{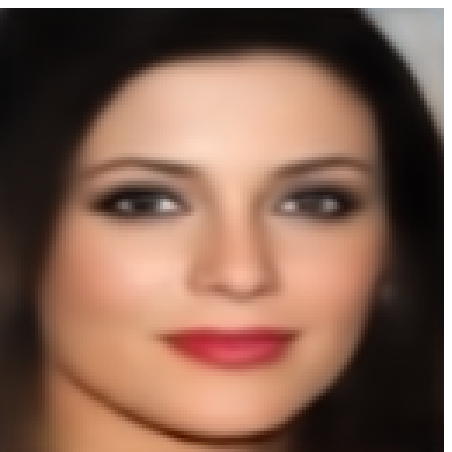}
\includegraphics[width=0.10\linewidth]{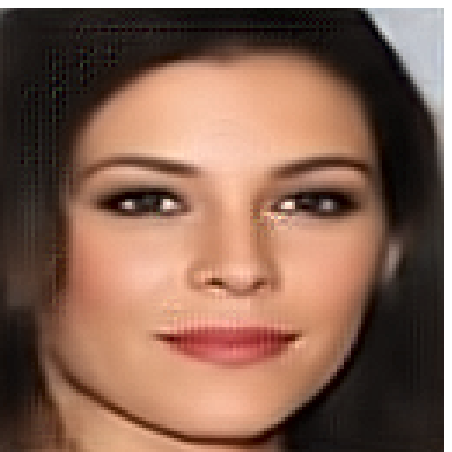}\\
\vspace{-1.2mm}
\subfigure[]{\label{fig:Desa}\scalebox{1}[1]{\includegraphics[width=0.10\linewidth]{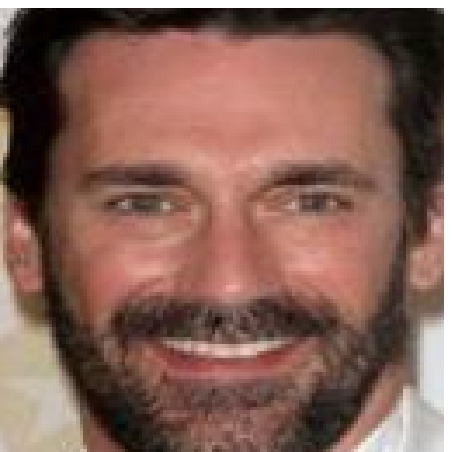}}}
\subfigure[]{\label{fig:Desb}\scalebox{1}[1]
{\includegraphics[width=0.10\linewidth]{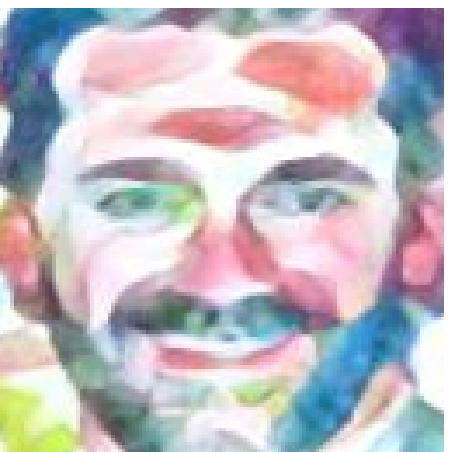}}}
\subfigure[]{\label{fig:Desc}\scalebox{1}[1]
{\includegraphics[width=0.10\linewidth]{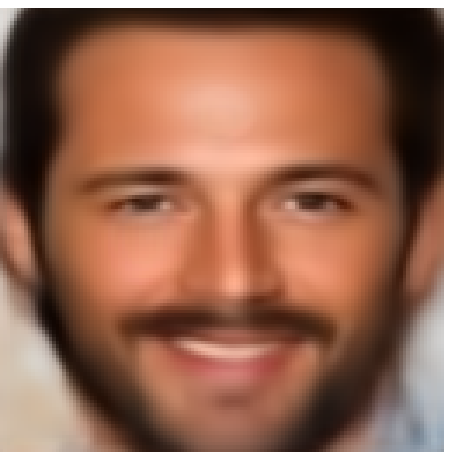}}}
\subfigure[]{\label{fig:Desd}\scalebox{1}[1]
{\includegraphics[width=0.10\linewidth]{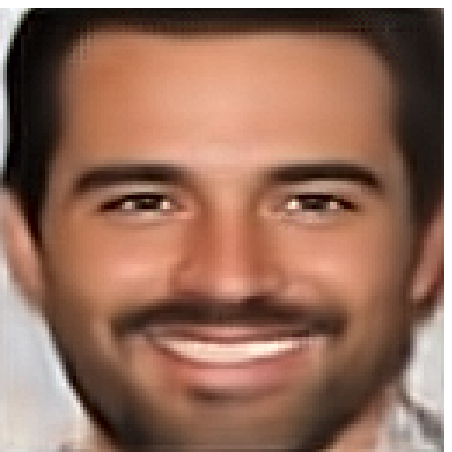}}}
\subfigure[]{\label{fig:Dese}\scalebox{1}[1]
{\includegraphics[width=0.10\linewidth]{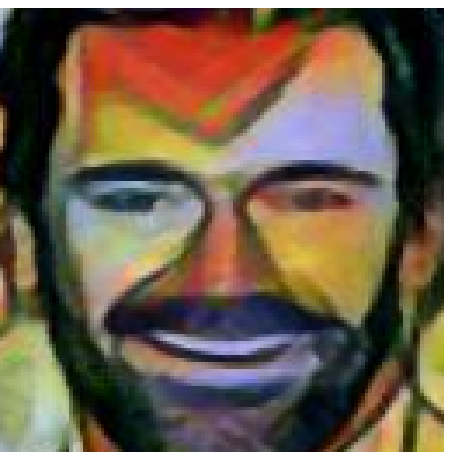}}}
\subfigure[]{\label{fig:Desf}\scalebox{1}[1]
{\includegraphics[width=0.10\linewidth]{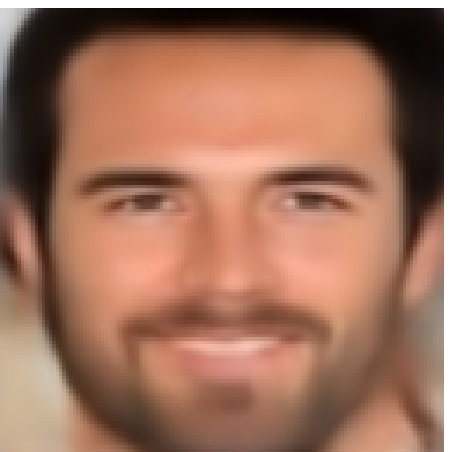}}}
\subfigure[]{\label{fig:Desg}\scalebox{1}[1]
{\includegraphics[width=0.10\linewidth]{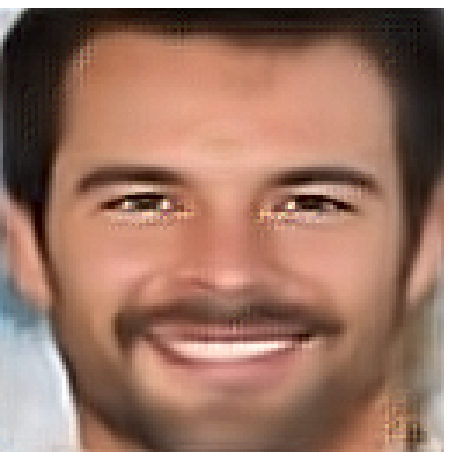}}}
\caption{Contribution of each FDNN part. (a) Ground-truth real face images. (b) Input portrait of \emph{Feathers} from training styles and (e) input portrait of \emph{la Muse} from unseen styles (from test dataset; not available in the training stage). (c, f) Destylization results without adversarial loss. (d, g) Our final results.}
\label{fig:DesEffect}
\end{figure*}

Our training strategy enforces the generated face $\hat{r}_i$ to be similar to its corresponding ground-truth $r_i$. Therefore, we employ a pixel-wise $\ell_2$ loss between $\hat{r}_i$ and $r_i$, and
we minimize the objective $Q(\mathcal{T})$ of SRN as follows:

\begin{equation}
\label{eqn:SRN}
\begin{split}
\min\limits_{\mathcal{T}}\;\;Q(\mathcal{T})\!&=\!\mathbb{E}_{(\hat{r}_i,r_i)\sim p(\hat{r},r)} \|\hat{r}_i - r_i\|_F^2\\
&=\!\mathbb{E}_{(s_i,r_i)\sim p(s,r)} \|G_{\mathcal{T}}(s_i) - r_i\|_F^2,
\end{split}
\end{equation}
where $\mathcal{T}$ indicates the parameters of SRN generator G, $p(s,r)$ represents the joint distribution of the SF and RF images in the training dataset and $p(\hat{r},r)$ represents the joint distribution of destylized and the ground-truth faces.

To achieve high-quality results, we force SRN to fool the discriminative supervising network that employs a binary classifier which task is to distinguish whether incoming image samples contain real or generated faces. Similar to the idea of~\cite{Goodfellow2014, Denton2015, radford2015unsupervised}, our goal is to make the discriminative network fail to distinguish generated faces from real ones. Hereby, we maximize the adversarial loss of the discriminative network $F(\mathcal{L})$ as follows:
\begin{equation}
\hspace{-1em}
\label{eqn:A}
\begin{split}
&\max\limits_{\mathcal{L}}\;\;F(\mathcal{L})\!=\!\mathbb{E}\left[\log D_\mathcal{L}(r_i) + \log(1-D_\mathcal{L}(\hat{r}_i))\right]\\
& \qquad\qquad=\!\mathbb{E}_{r_i\sim p(r)}[\log D_\mathcal{L}(r_i)]\!+\!\mathbb{E}_{\hat{r}_i\sim p(\hat{r}))}[\log(1\!-\!D_\mathcal{L}(\hat{r}_i))], 
\end{split}\!\!\!
\end{equation}
where $\mathcal{L}$ represents the parameters of the discriminative network $D$, $p(r)$ and $p(\hat{r})$ indicate the distributions corresponding to the real and the generated faces, respectively, and $D_\mathcal{L}(r_i)$ and $D_\mathcal{L}(\hat{r}_i)$ are the outputs of network $D$. Since the loss $F$ is back-propagated to update not only the parameters $\mathcal{L}$ but also $\mathcal{T}$, we also minimize the objective function $Q_f(\mathcal{T})$ of SRN: 
\begin{equation}
\begin{split}
\min\limits_{\mathcal{T}}\;\;\;Q_f(\mathcal{T})&\!=\!\mathbb{E}_{(s_i,r_i)\sim p(s,r)} \|G_{\mathcal{T}}(s_i) - r_i\|_F^2\\
&\!+\!\lambda \mathbb{E}_{s_i\sim p(s))}[\log\!D_\mathcal{L}(G_{\mathcal{T}}(s_i))],
\end{split}
\end{equation}
where scalar $\lambda$ is a trade-off between supervising the generator by the ground-truth data vs. the discriminator supervision, respectively.

Since each layer in our FDNN is differentiable, we employ the Root Mean Square Propagation (RMSprop)~\cite{Hinton} to update $\mathcal{T}$ and $\mathcal{L}$. In order to maximize the adversarial loss $F$, the stochastic gradient ascent is used to update $\mathcal{L}$:
\begin{equation}
\begin{split}
\Delta^{i+1} &= \beta \Delta ^{i} + (1-\beta)(\frac{\partial F}{\partial \mathcal{L}})^2,\\
\mathcal{L}^{i+1} &= \mathcal{L}^{i} + \alpha \frac{\partial F}{\partial \mathcal{L}} \frac{1}{\sqrt{\Delta^{i+1}+\epsilon}},
\end{split}
\end{equation}
where $\alpha$ and $\beta$ represent the learning and the decay rate respectively, $i$ is the iteration index, $\Delta$ is an auxiliary variable, and $\epsilon$ is set to $10^{-8}$ to avoid division by zero. For SRN, both losses $Q$ and $F$ are used to update $\mathcal{T}$ by the stochastic gradient descent:
\begin{equation}
\begin{split}
\Delta^{i+1} &= \beta \Delta ^{i} + (1-\beta)(\frac{\partial Q_f}{\partial \mathcal{T}})^2,\\
\mathcal{T}^{i+1} &= \mathcal{T}^{i} - \alpha (\frac{\partial Q_f}{\partial \mathcal{T}})\frac{1}{\sqrt{\Delta^{i+1}+\epsilon}},
\end{split}
\end{equation} 

We set $\lambda=0.01$ to limit supervision of the generator by the discriminator and allow appearance-based learning from the ground-truth image pairs. 
As the iterations progress, the output faces will resemble the real faces more.  Therefore, we gradually reduce the impact of the discriminative network by decreasing $\lambda$,
\begin{equation}
\label{eqn:lambda}
\lambda^{n} = \max\{\lambda\cdot 0.995^n, \lambda/2 \},
\end{equation}
where $n$ is the index of the epochs. Eqn.~\ref{eqn:lambda} not only increases the impact of the appearance similarity term but also preserves the class-specific discriminative information in the training phase.

\subsection{Implementation Details}
Similar to ~\cite{Goodfellow2014,radford2015unsupervised}, we employ batch normalization after the convolutional and deconvolutional layers of SRN except for the last deconvolutional layers. We also use leaky rectified linear units (leakyReLU) with a negative slope $0.2$ as non-linear activation functions. 
For training, the learning rate $\alpha$ is set to $0.001$ and multiplied by $0.99$ after each epoch, and the decay rate is set to $0.01$. 
The discriminative network is only employed in the training phase. In the testing phase, we feed a stylized face image into the SRN to obtain its realistic version. 

\begin{figure}
\begin{minipage}{0.19\linewidth}
\centering
\subfigure[]{\label{fig:fig3a}\scalebox{1}[1]{\includegraphics[width=1\linewidth]{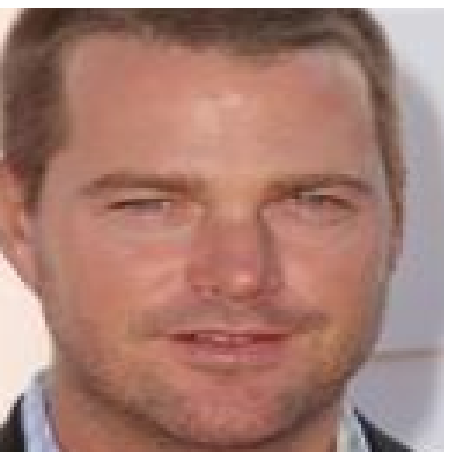}}}
\end{minipage}
\begin{minipage}{0.8\linewidth}
\centering
\subfigure[]{\label{fig:fig3b}\scalebox{1}[1]{\includegraphics[width=0.23\linewidth]{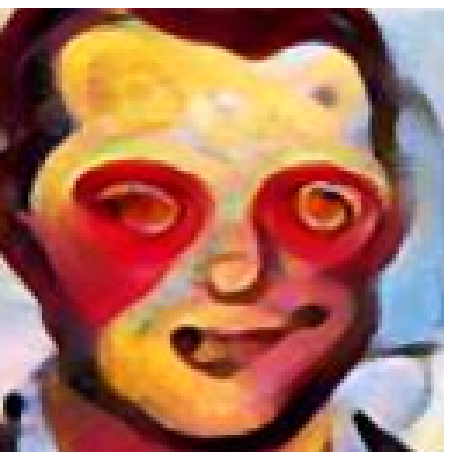}}}
\subfigure[]{\label{fig:fig3c}\scalebox{1}[1]{\includegraphics[width=0.23\linewidth]{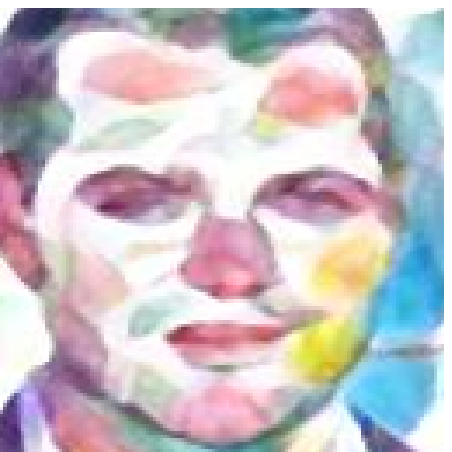}}}
\subfigure[]{\label{fig:fig3d}\scalebox{1}[1]{\includegraphics[width=0.23\linewidth]{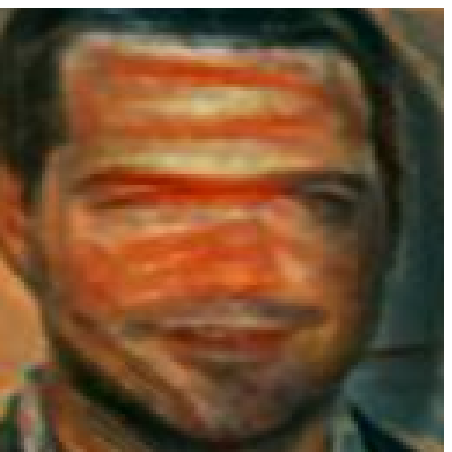}}}
\subfigure[]{\label{fig:fig3e}\scalebox{1}[1]{\includegraphics[width=0.23\linewidth]{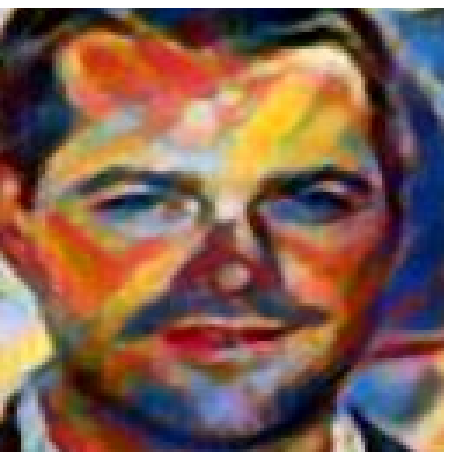}}}\\\vspace{-0.8em}
\subfigure[]{\label{fig:fig3f}\scalebox{1}[1]{\includegraphics[width=0.23\linewidth]{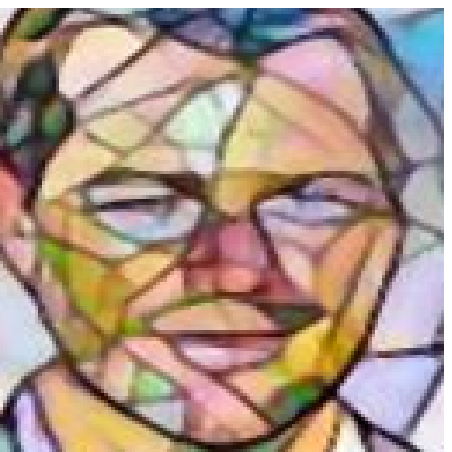}}}
\subfigure[]{\label{fig:fig3d}\scalebox{1}[1]{\includegraphics[width=0.23\linewidth]{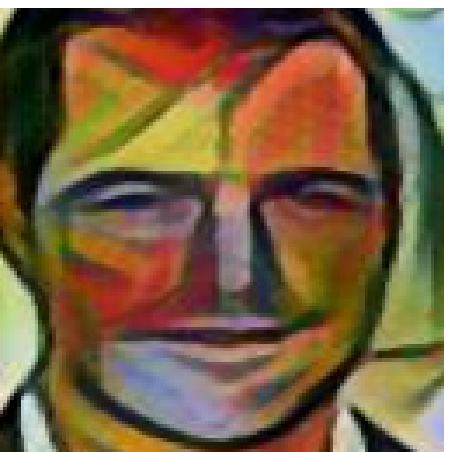}}}
\subfigure[]{\label{fig:fig3e}\scalebox{1}[1]{\includegraphics[width=0.23\linewidth]{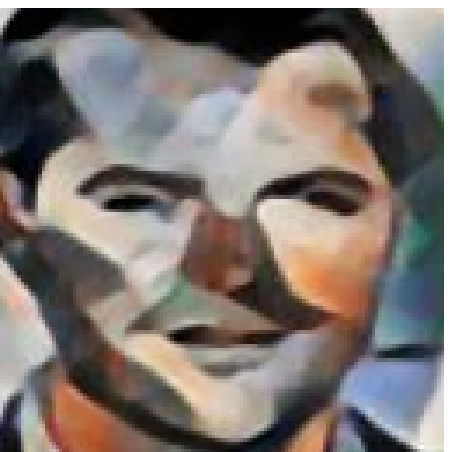}}}
\subfigure[]{\label{fig:fig3g}\scalebox{1}[1]{\includegraphics[width=0.23\linewidth]{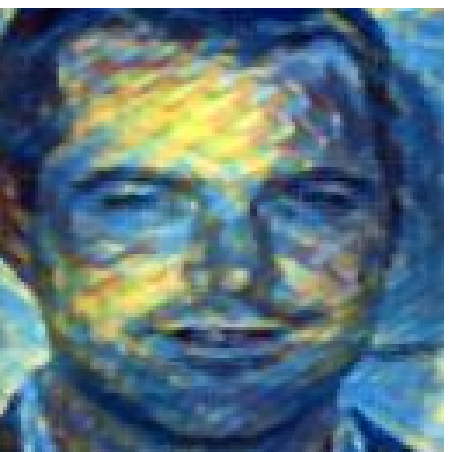}}}
\end{minipage}
\caption{Illustration of the synthesized dataset. (a) Original real face image. (b)-(d) The synthesized stylized faces of (a) form \emph{Candy, Feathers} and \emph{Scream} which have been used for training our network. (e)-(i) The synthesized stylized faces of (a) form \emph{Composition VII, Mosaic, la Muse, Udnie} and $Starry$ styles which have not been used for training.}
\label{fig:dataset}
\end{figure}

\section{Synthesized Dataset}
\label{Sec:Data}

Training of a deep neural network requires a large number of samples to prevent models from overfitting to the training data. The publicly available large-scale face datasets~\cite{LFWTech, Liu2015faceattributes} only provide faces in the wild but not pairs of real images of faces and their stylizations. Therefore, we opt to generate a large number of stylized faces from the corresponding real face images in eight distinct styles: \emph{Starry Night, la Muse, Composition VII, Scream, Candy, Feathers, Mosaic} and \emph{Udnie}. To generate such a dataset, there are a number of alternative feed-forward approaches available \cite{ulyanov2016texture,ulyanov2016instance,johnson2016perceptual}. We choose the recent feed-forward style transfer model \cite{johnson2016perceptual}.

We firstly select at random 10K images of cropped real faces (within $\pm 30^{\circ}$ orientation) from the CelebA~\cite{Liu2015faceattributes} dataset for training and 1K images for testing, and then resize them to 128$\times$128 pixels. We use 10K training images as our real ground-truth faces $r_i$. To generate three different portraits of each face, we retrain the style transfer model \cite{johnson2016perceptual} for \emph{Scream, Candy} and \emph{Feathers} styles separately. Finally, we obtain 30K SF/RF pairs for training our network. We also use 1K test real faces to generate 8K SF/RF face pairs from eight different styles (each test face corresponds to eight distinct styles) for testing our network. Figure~\ref{fig:dataset} shows the stylized samples that are generated from a single real image containing a face (Fig.~\ref{fig:fig3a}).

\section{Experiments}
We compare our method qualitatively and quantitatively against four different state-of-the-art methods. As explained in Sec.~\ref{Sec:Data}, we gather 30K SF/RF face pairs from three styles as a training dataset and 8K SF/RF pairs faces generated from different eight styles for testing. In all the cases, the ground-truth real faces and the corresponding stylized faces do not overlap in the training and testing datasets. Since our method is feed-forward and no optimization is required at test time. Our method cost 10 ms for a 128-by-128 image.
 
\begin{figure*}[t]
\centering
\includegraphics[width=0.11\linewidth]{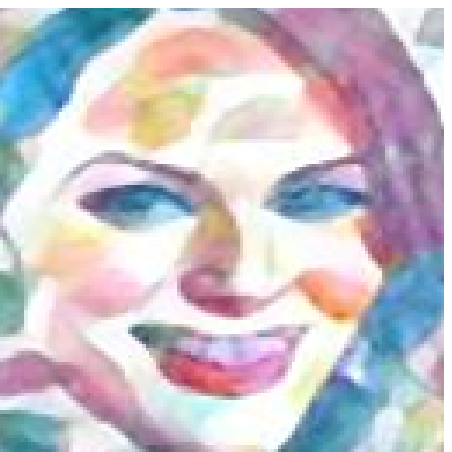}
\includegraphics[width=0.11\linewidth]{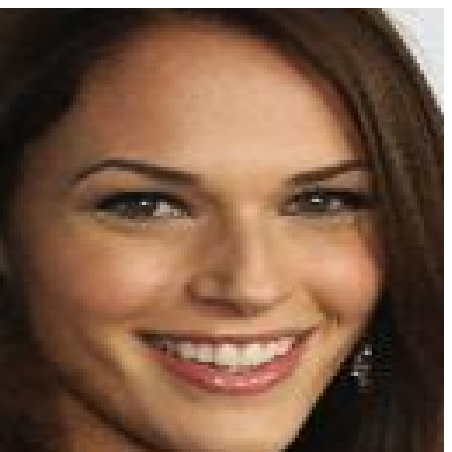}
\includegraphics[width=0.11\linewidth]{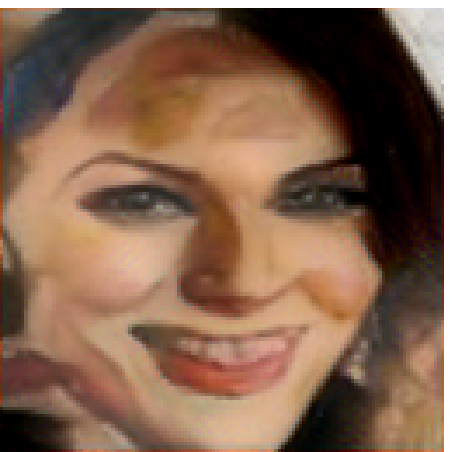}
\includegraphics[width=0.11\linewidth]{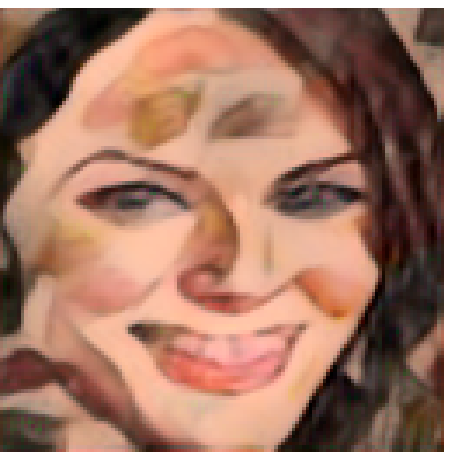}
\includegraphics[width=0.11\linewidth]{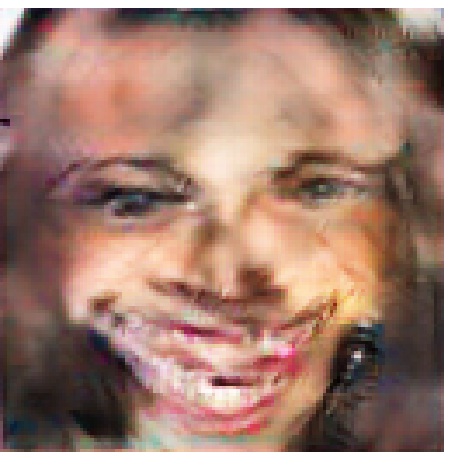}
\includegraphics[width=0.11\linewidth]{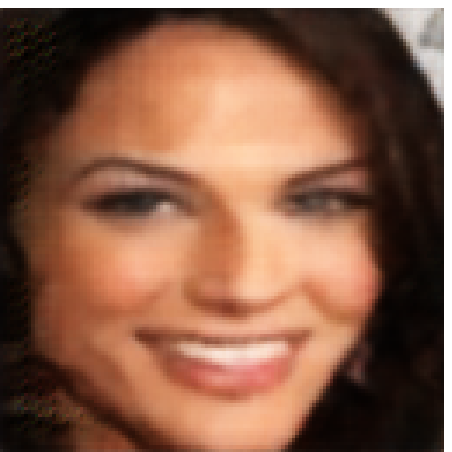}
\includegraphics[width=0.11\linewidth]{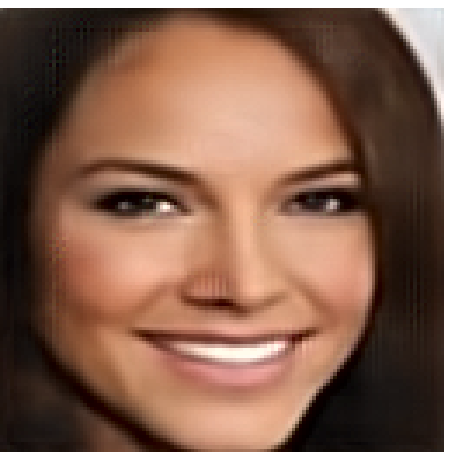}\\\vspace{0.1em}
\includegraphics[width=0.11\linewidth]{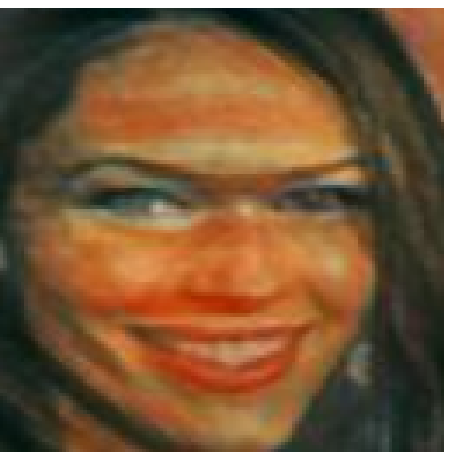}
\includegraphics[width=0.11\linewidth]{figs/015248.eps}
\includegraphics[width=0.11\linewidth]{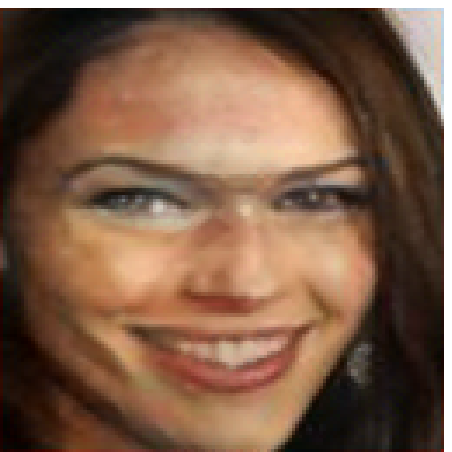}
\includegraphics[width=0.11\linewidth]{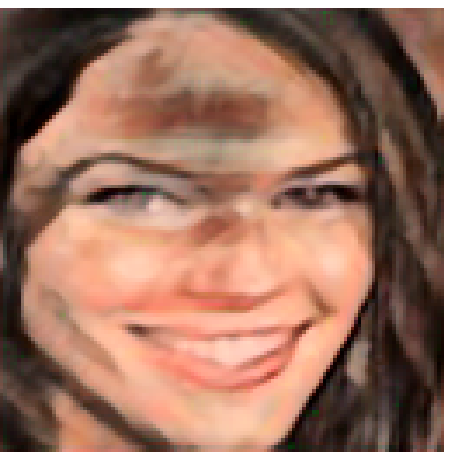}
\includegraphics[width=0.11\linewidth]{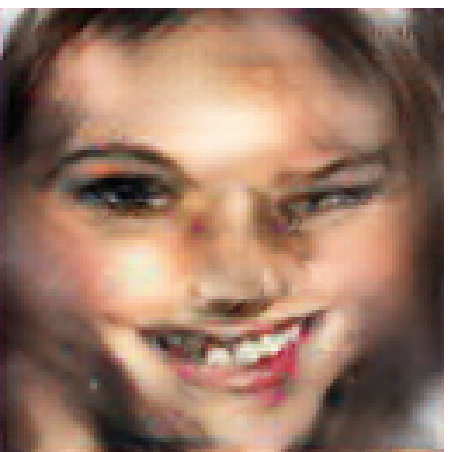}
\includegraphics[width=0.11\linewidth]{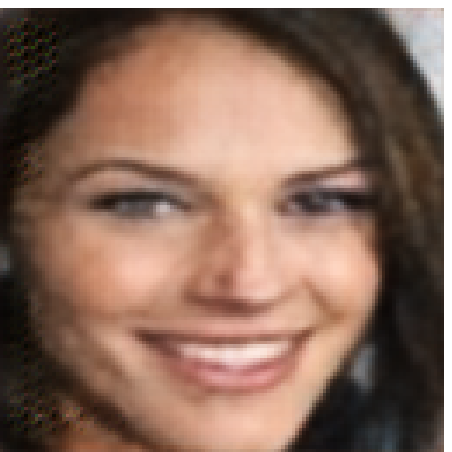}
\includegraphics[width=0.11\linewidth]{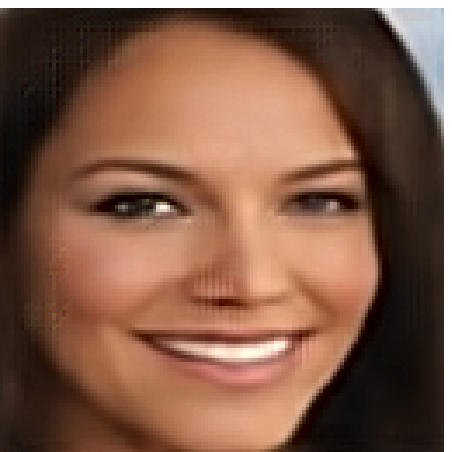}\\\vspace{0.1em}
\includegraphics[width=0.11\linewidth]{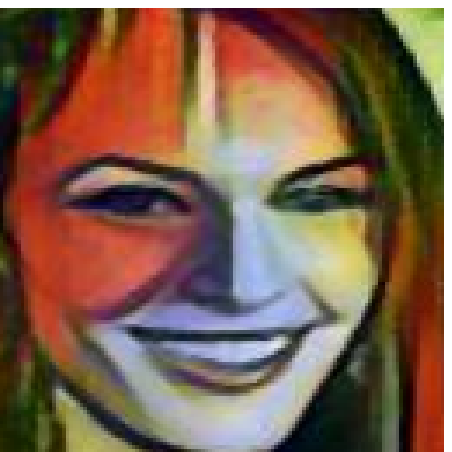}
\includegraphics[width=0.11\linewidth]{figs/015248.eps}
\includegraphics[width=0.11\linewidth]{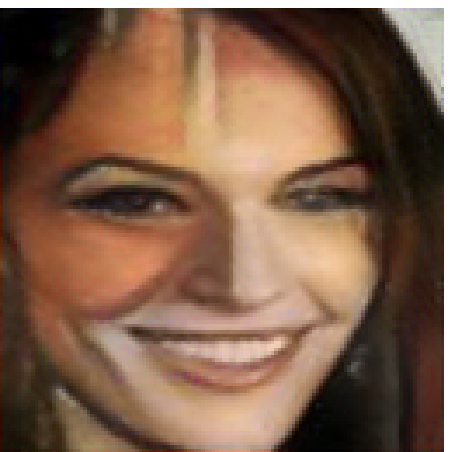}
\includegraphics[width=0.11\linewidth]{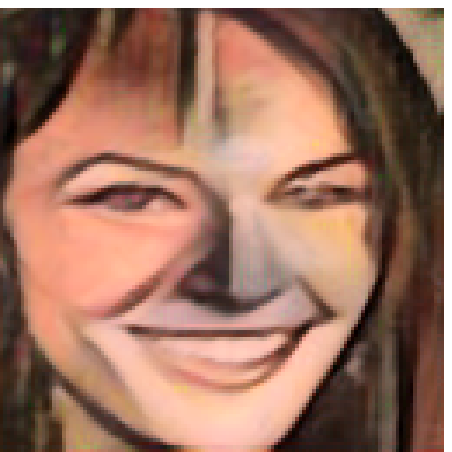}
\includegraphics[width=0.11\linewidth]{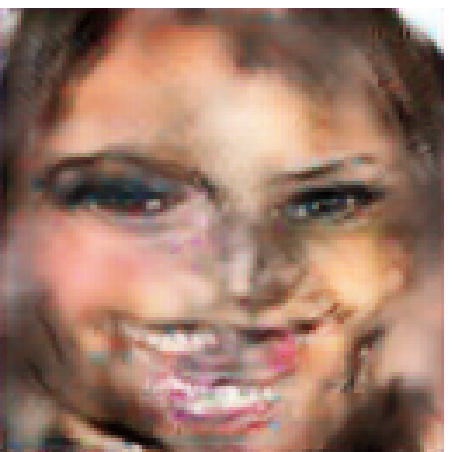}
\includegraphics[width=0.11\linewidth]{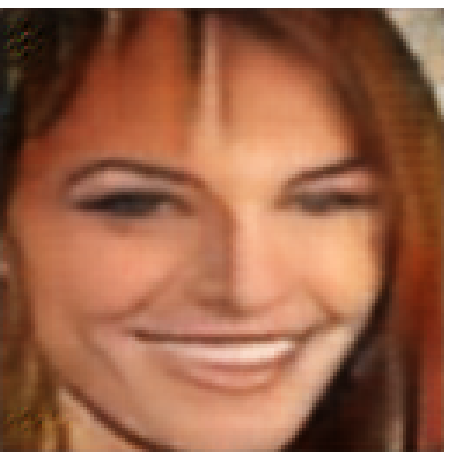}
\includegraphics[width=0.11\linewidth]{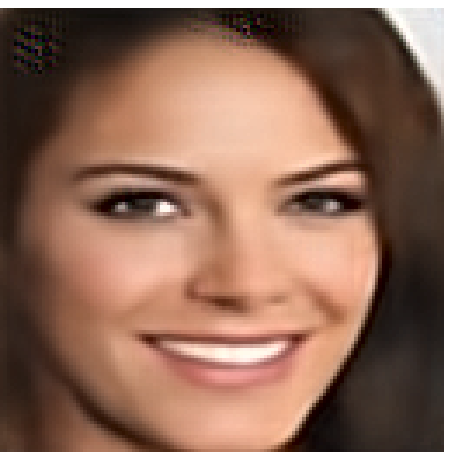}\\\vspace{0.1em}
\includegraphics[width=0.11\linewidth]{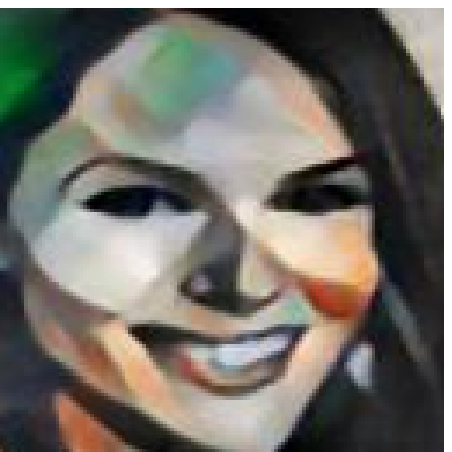}
\includegraphics[width=0.11\linewidth]{figs/015248.eps}
\includegraphics[width=0.11\linewidth]{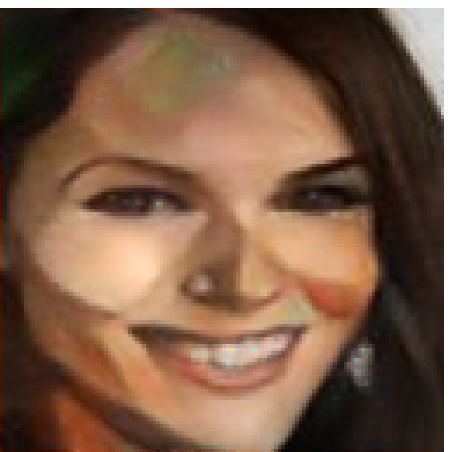}
\includegraphics[width=0.11\linewidth]{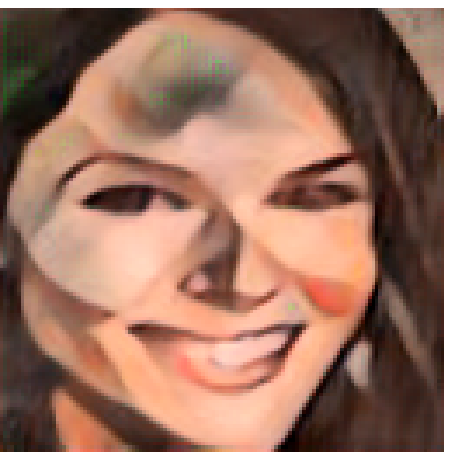}
\includegraphics[width=0.11\linewidth]{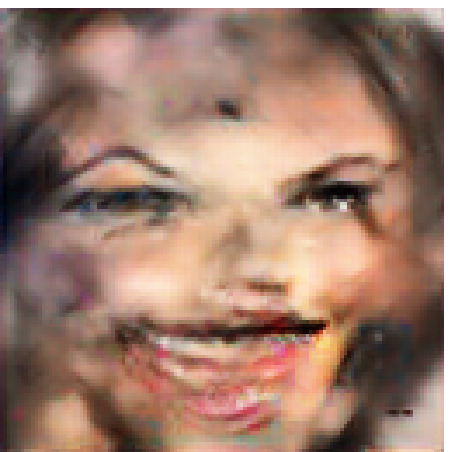}
\includegraphics[width=0.11\linewidth]{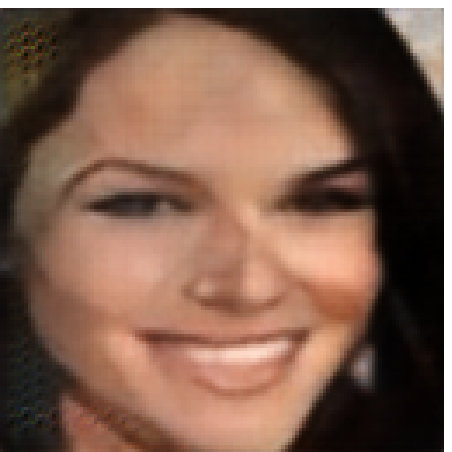}
\includegraphics[width=0.11\linewidth]{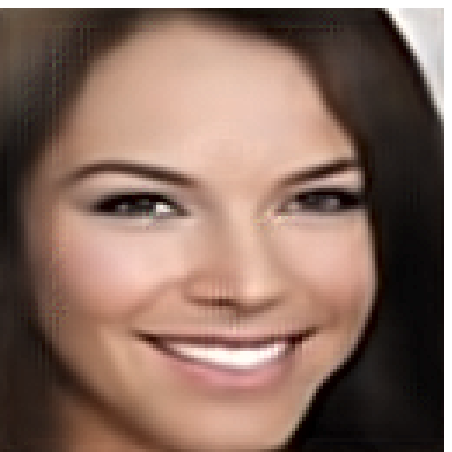}\\
\vspace{-1.2mm}
\subfigure[SF]{\label{fig:cmp1a}\scalebox{1}[1]{\includegraphics[width=0.11\linewidth]{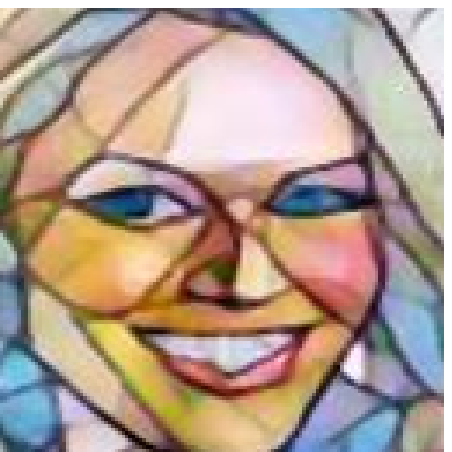}}}
\subfigure[Groundtruth]{\label{fig:cmp1b}\scalebox{1}[1]
{\includegraphics[width=0.11\linewidth]{figs/015248.eps}}}
\subfigure[Gatys~\cite{gatys2016image}]{\label{fig:cmp1c}\scalebox{1}[1]
{\includegraphics[width=0.11\linewidth]{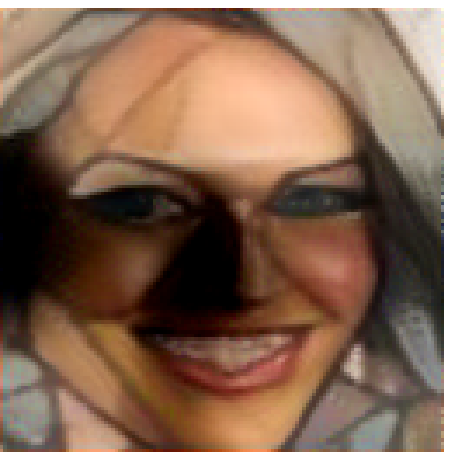}}}
\subfigure[Johnson~\cite{johnson2016perceptual}]{\label{fig:cmp1d}\scalebox{1}[1]
{\includegraphics[width=0.11\linewidth]{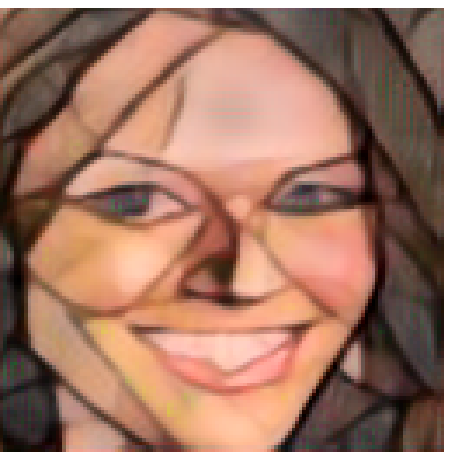}}}
\subfigure[MGAN~\cite{li2016precomputed}]{\label{fig:cmp1e}\scalebox{1}[1]
{\includegraphics[width=0.11\linewidth]{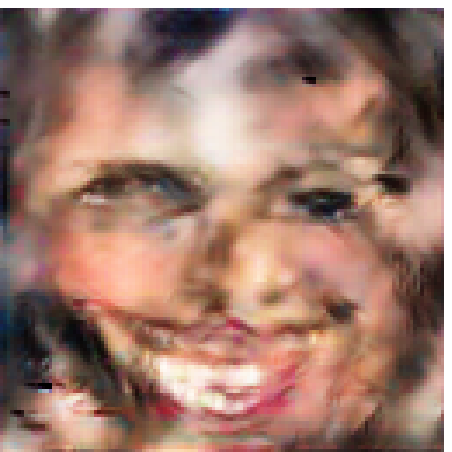}}}
\subfigure[pix2pix~\cite{isola2016image}]{\label{fig:cmp1f}\scalebox{1}[1]
{\includegraphics[width=0.11\linewidth]{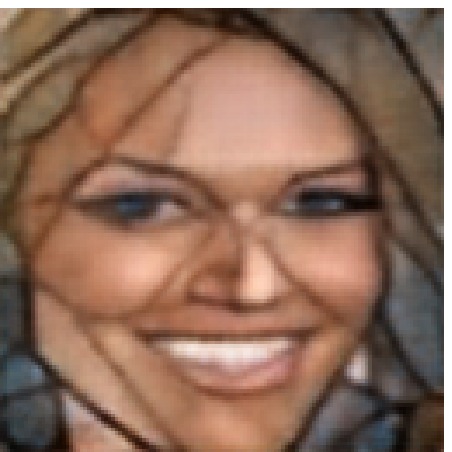}}}
\subfigure[Ours]{\label{fig:cmp1g}\scalebox{1}[1]
{\includegraphics[width=0.11\linewidth]{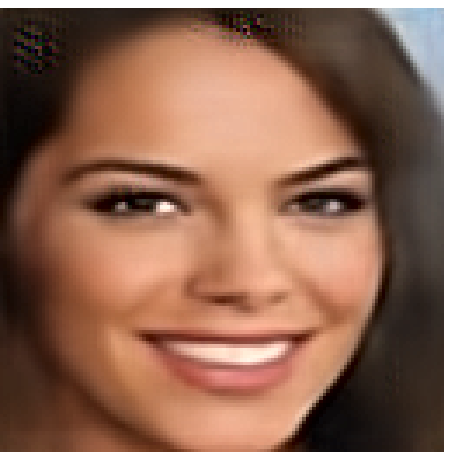}}}
\caption{Results of the state-of-the-art methods for face destylization. (a) Input portraits of \emph{Feathers, Scream} from seen styles as well as \emph{la Muse, Udnie} and \emph{Mosaic} from unseen styles (from test dataset; not available to the algorithm during training) (b) Ground-truth images of real faces. 
 }
\label{fig:cmp1}
\vspace{-0.5cm}
\end{figure*}

\begin{figure*}[t]
\centering
\includegraphics[width=0.11\linewidth]{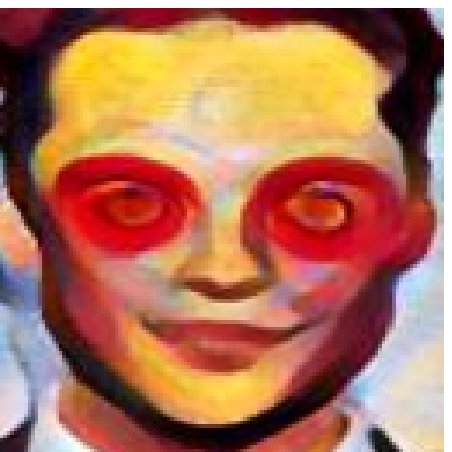}
\includegraphics[width=0.11\linewidth]{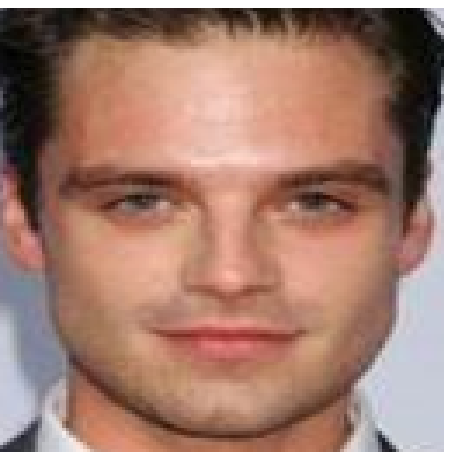}
\includegraphics[width=0.11\linewidth]{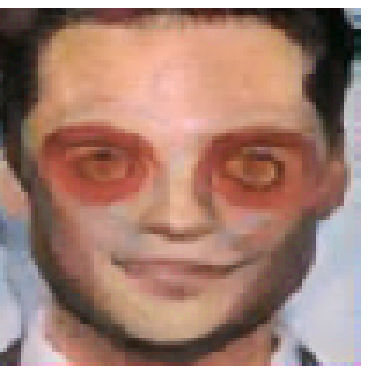}
\includegraphics[width=0.11\linewidth]{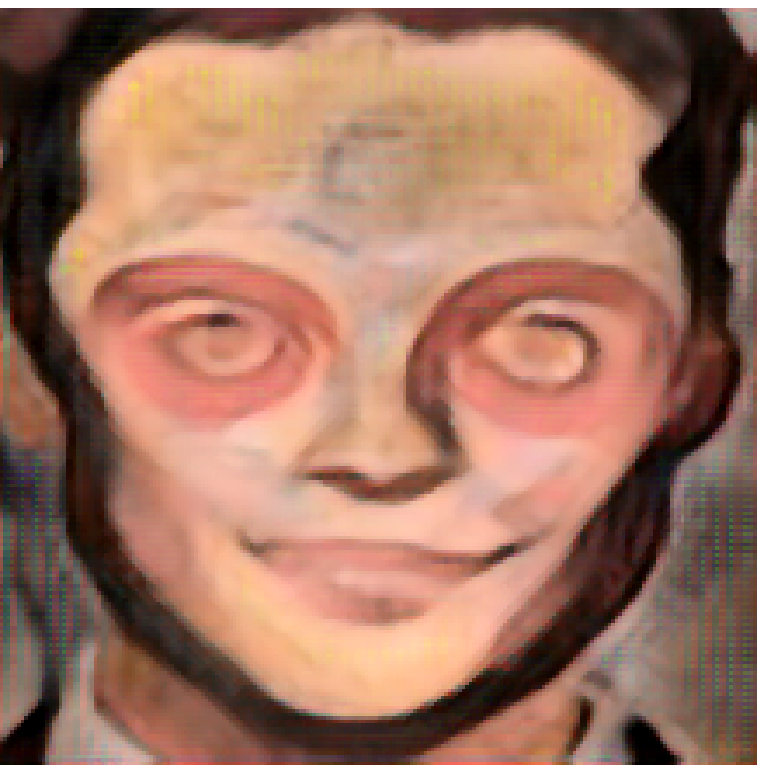}
\includegraphics[width=0.11\linewidth]{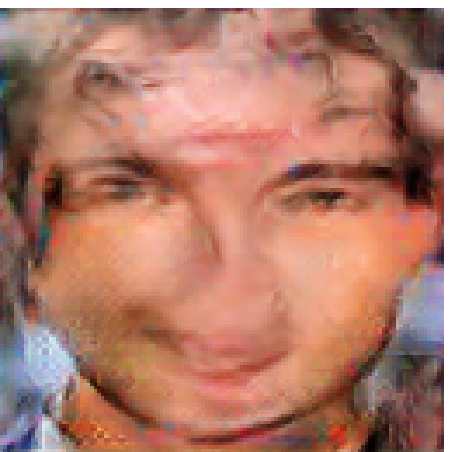}
\includegraphics[width=0.11\linewidth]{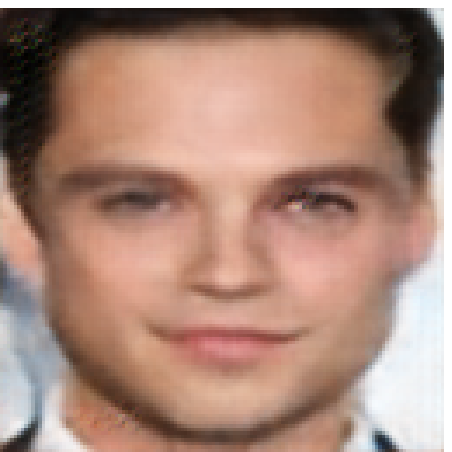}
\includegraphics[width=0.11\linewidth]{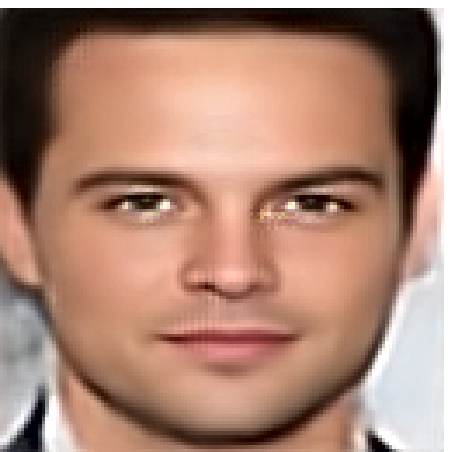}\\\vspace{0.1em}
\includegraphics[width=0.11\linewidth]{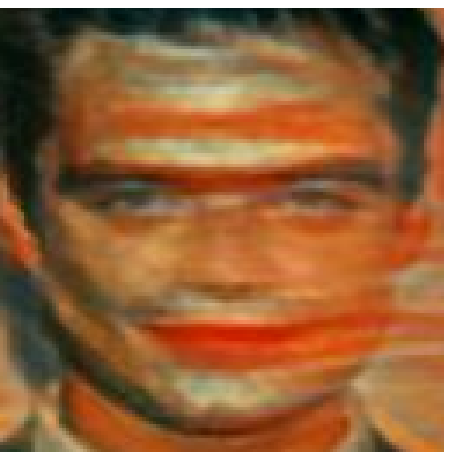}
\includegraphics[width=0.11\linewidth]{figs/013944.eps}
\includegraphics[width=0.11\linewidth]{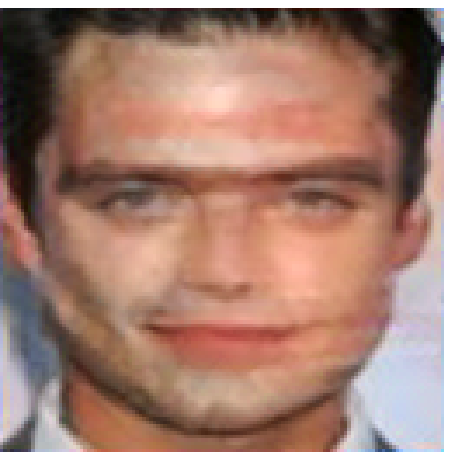}
\includegraphics[width=0.11\linewidth]{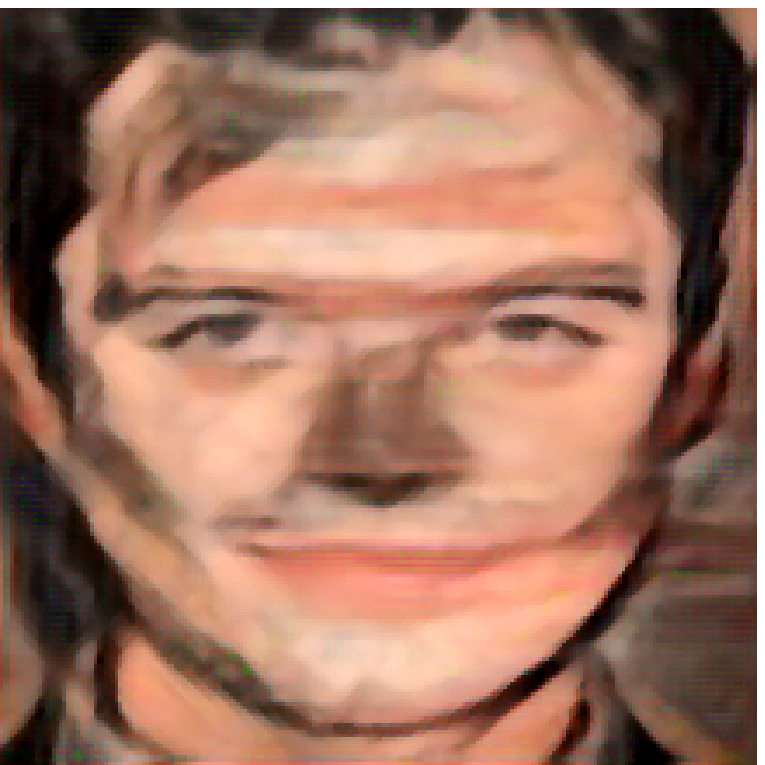}
\includegraphics[width=0.11\linewidth]{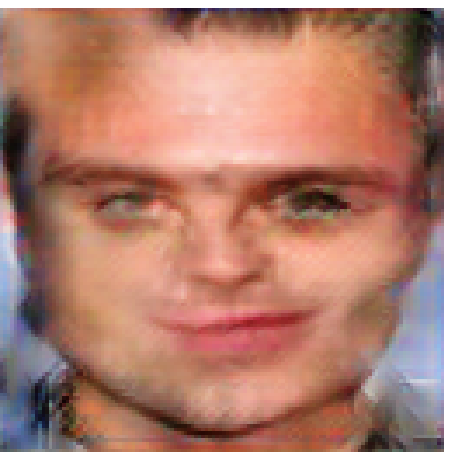}
\includegraphics[width=0.11\linewidth]{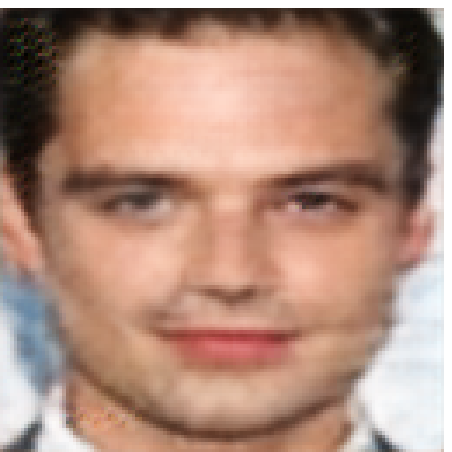}
\includegraphics[width=0.11\linewidth]{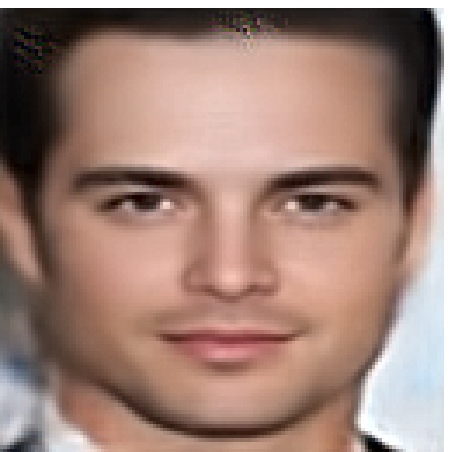}\\\vspace{0.1em}
\includegraphics[width=0.11\linewidth]{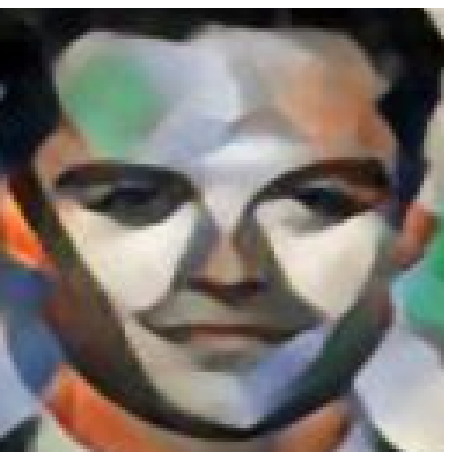}
\includegraphics[width=0.11\linewidth]{figs/013944.eps}
\includegraphics[width=0.11\linewidth]{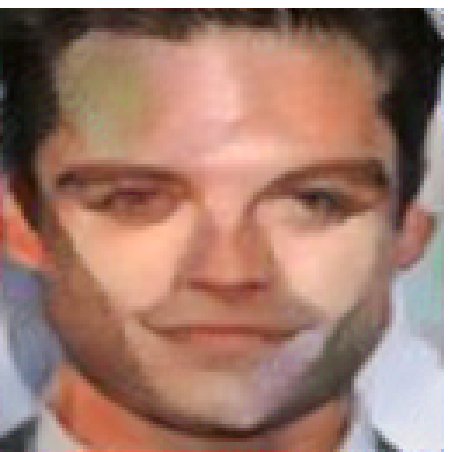}
\includegraphics[width=0.11\linewidth]{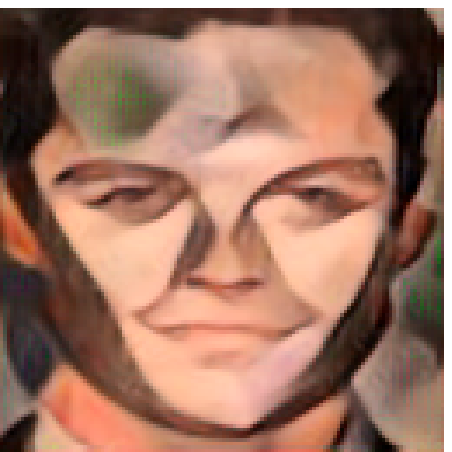}
\includegraphics[width=0.11\linewidth]{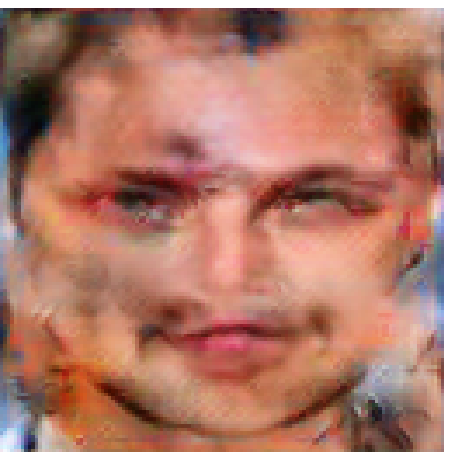}
\includegraphics[width=0.11\linewidth]{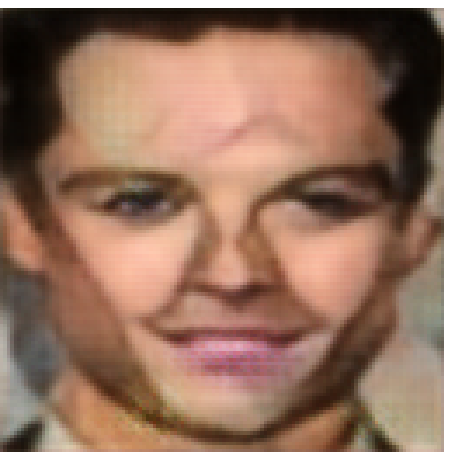}
\includegraphics[width=0.11\linewidth]{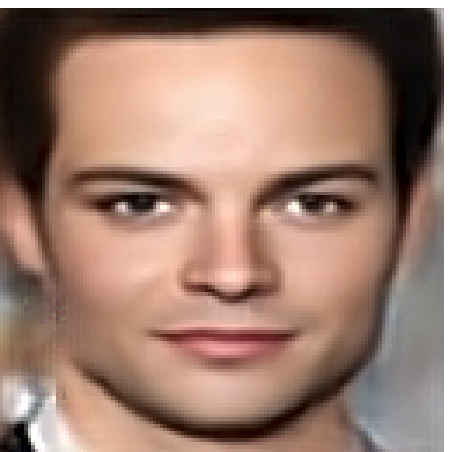}\\\vspace{0.1em}
\includegraphics[width=0.11\linewidth]{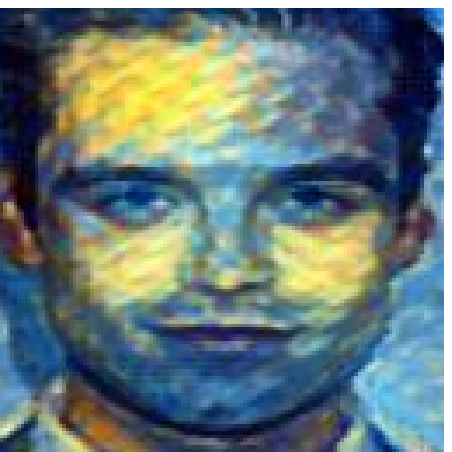}
\includegraphics[width=0.11\linewidth]{figs/013944.eps}
\includegraphics[width=0.11\linewidth]{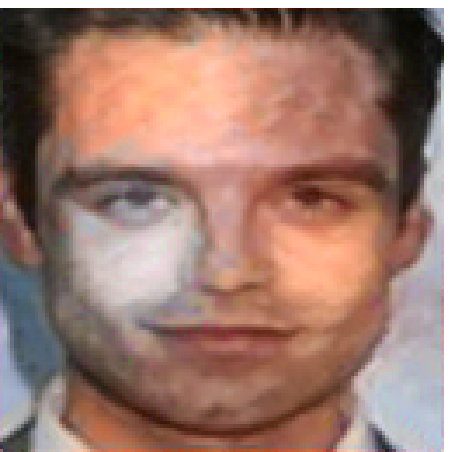}
\includegraphics[width=0.11\linewidth]{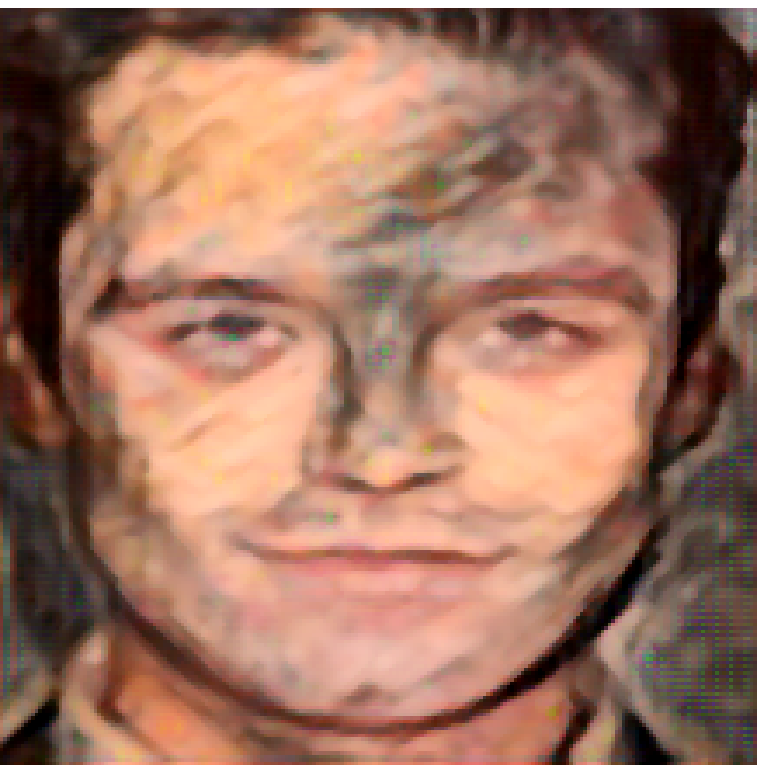}
\includegraphics[width=0.11\linewidth]{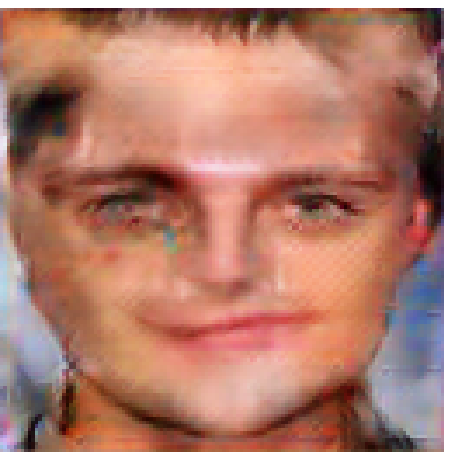}
\includegraphics[width=0.11\linewidth]{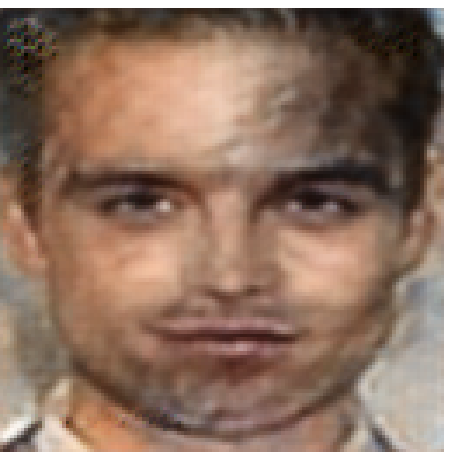}
\includegraphics[width=0.11\linewidth]{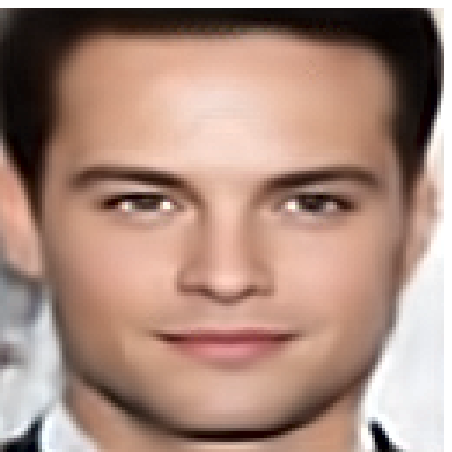}\\
\vspace{-1.2mm}
\subfigure[SF]{\label{fig:cmp2a}\scalebox{1}[1]{\includegraphics[width=0.11\linewidth]{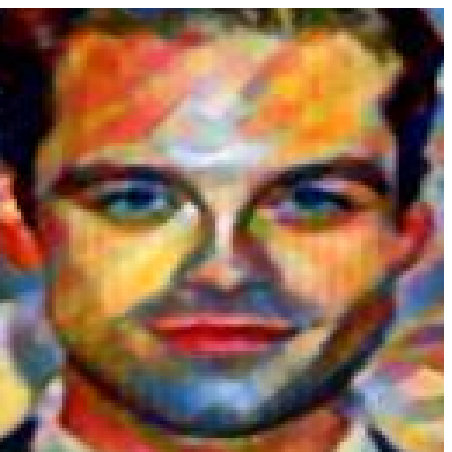}}}
\subfigure[Groundtruth]{\label{fig:cmp2b}\scalebox{1}[1]
{\includegraphics[width=0.11\linewidth]{figs/013944.eps}}}
\subfigure[Gatys~\cite{gatys2016image}]{\label{fig:cmp2c}\scalebox{1}[1]
{\includegraphics[width=0.11\linewidth]{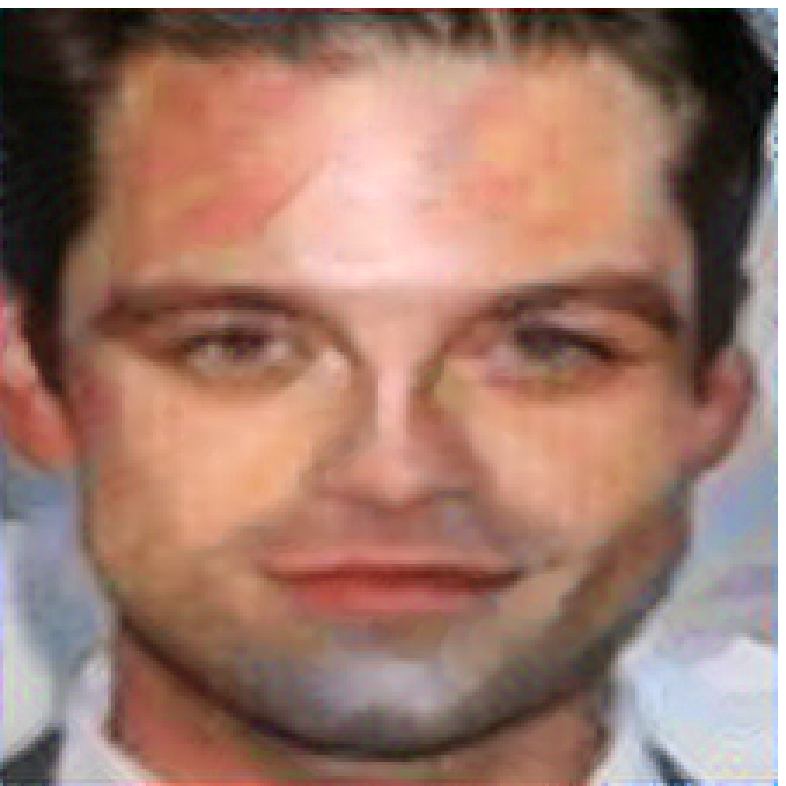}}}
\subfigure[Johnson~\cite{johnson2016perceptual}]{\label{fig:cmp2d}\scalebox{1}[1]
{\includegraphics[width=0.11\linewidth]{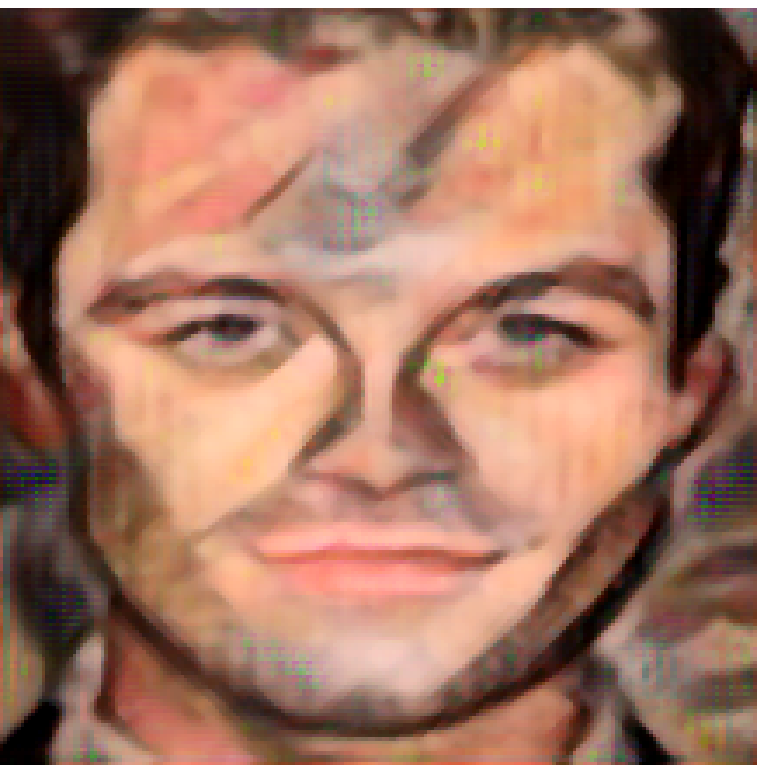}}}
\subfigure[MGAN~\cite{li2016precomputed}]{\label{fig:cmp2e}\scalebox{1}[1]
{\includegraphics[width=0.11\linewidth]{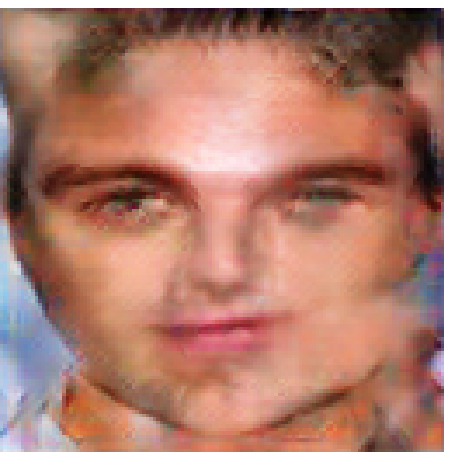}}}
\subfigure[pix2pix~\cite{isola2016image}]{\label{fig:cmp2f}\scalebox{1}[1]
{\includegraphics[width=0.11\linewidth]{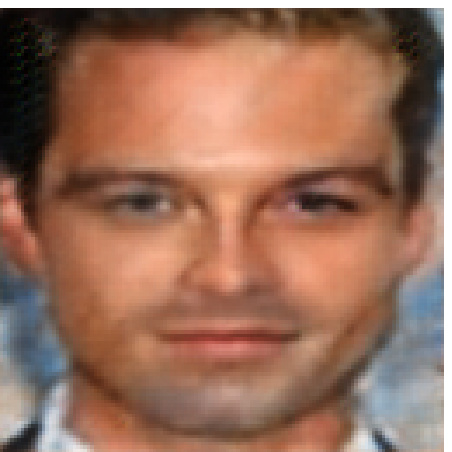}}}
\subfigure[Ours]{\label{fig:cmp2g}\scalebox{1}[1]
{\includegraphics[width=0.11\linewidth]{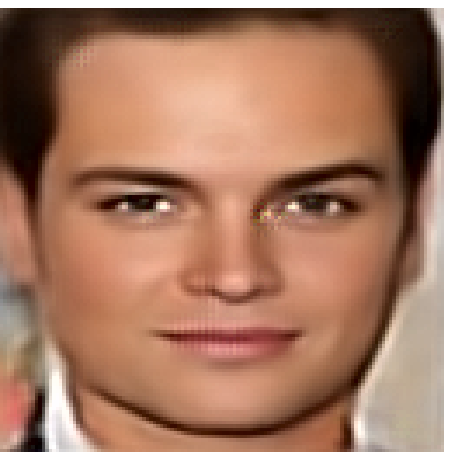}}}
\caption{Result of the state-of-the-art methods for face destylization. (a) Input portraits of \emph{Candy} and \emph{Scream} from seen styles as well as \emph{la Muse, starry Night} and \emph{Mosaic} from unseen styles (from test dataset; not available to the algorithm during training) (b) Ground-truth images of real faces.}
\label{fig:cmp2}
\vspace{-0.5cm}
\end{figure*}

\subsection{Qualitative Evaluation}
\textbf{Comparison to the state of the art.} Firstly, we note that the test stylized face images were not used for training of our model. The resolution of stylized and destylized output faces in this study is $128\times128$ pixels. We compare our approach against four different approaches as detailed below.

We compare our work against \cite{gatys2016image}, an image-optimization based style transfer method free of the training stage. To generate real faces, this network aims to preserve the contents of a portrait and the corresponding photo-realistic face. The network fails to produce appealing results as shown in Fig.~\ref{fig:cmp1c} and Fig.~\ref{fig:cmp2c}. This method captures the correlations in feature maps of style and synthesized images by Gram matrices and discards the spatial arrangement at the pixel level. 

We also use a feed-forward approach \cite{johnson2016perceptual} for destylization. Due to the Gram matrix, this method also produces distorted facial details. As shown in the first row of Fig.~\ref{fig:cmp1d}, the edges of the face were blurred and the color of the face is not consistent. From the first row of Fig.~\ref{fig:cmp2d}, one can see that the style overlapping with the eyes was not fully removed. Thus, their network fails to restore authentic looking eyes.

Li and Wand \cite{li2016precomputed} propose a patch-based style transfer method, known as Markovian GAN. We use their network for destylization and apply their standard protocols. As such a method is trained with stylized face patches, it cannot capture the global structure of facial images. As seen in Fig.~\ref{fig:cmp1e} and Fig.~\ref{fig:cmp2e}, the facial color consistency cannot be preserved either. In contrast, our method produces highly-consistent facial colors and captures the global structure of faces well.

Isola \emph{et al.} \cite{isola2016image} present a general image-to-image translation method, known as pix2pix. It employs the architecture "Unet" for the generator network. A convolutional patch based neural network is trained to discriminate between image patches extracted from real and generated faces.
In addition, the low-level features from the bottom layers of Unet also participate in generating faces. These low-level features corrupt the destylized images and result in poor removal of styles in the images e.g., for unseen styles.
As shown in Fig.~\ref{fig:cmp1f} and Fig.~\ref{fig:cmp2f}, while pix2pix can produce acceptable results for seen styles, it fails to remove previously unseen styles.
As shown in the fourth row of Fig.~\ref{fig:cmp2f}, obvious artifacts appear in the generated face of an unseen style. 

Our destylized results exhibit higher fidelity w.r.t. the real faces, better consistency in colors and can even preserve the identity of the subject, as shown in Fig.~\ref{fig:cmp1g} and Fig.~\ref{fig:cmp2g}. 

\begin{table}\renewcommand{\arraystretch}{1.23}
\begin{center}
\caption{Comparison of physical (PSNR) and perceptual (SSIM) quality measures for the entire test dataset.}
\begin{tabular}{|c|c|c|c|c|}
\hline
\multirow{2}{*}{Method} & \multicolumn{2}{c|}{Seen Styles} & \multicolumn{2}{c|}{Unseen Styles} \\
\cline{2-5}
& PSNR & SSIM & PSNR & SSIM \\
\hline
Gatys ~\cite{gatys2016image} & 22.6792 & 0.8656 & 20.2320 & 0.8493\\ 
Johnson~\cite{johnson2016perceptual} & 22.8481 & 0.8745 & 21.2184 & 0.8632 \\
MGAN~\cite{li2016precomputed} & 19.5254 & 0.8548 & 17.2645 &  0.8270 \\
pix2pix~\cite{isola2016image} & 22.9893 & 0.8871 & 21.6316 & 0.8860 \\
\hline
Ours & {\bf 23.2086} & {\bf 0.9087} & {\bf 22.4430} & {\bf 0.9015}\\
\hline
\end{tabular}
\label{tab1}
\end{center}
\vspace{-0.3cm}
\end{table}

\subsection{Quantitative Evaluation}
\noindent{\textbf{Face Reconstruction. }}
In Tab. \ref{tab1}, we report the reconstruction performance  measured on the entire test dataset for each approach. We use the average Peak Signal to Noise Ratio (PSNR) and Structural Similarity (SSIM)~\cite{wang2004image} scores for which higher scores indicate better results. 

We report performance of destylization algorithms for two scenarios: seen and unseen styles. For the seen styles, results of the state-of-the-art style transfer methods are shown in the first and second rows of Fig.~\ref{fig:cmp1} and Fig.~\ref{fig:cmp2}. 
For the destylization of portraits of unseen styles, we demonstrate results in the third, fourth and fifth rows of Fig.~\ref{fig:cmp1} and Fig.~\ref{fig:cmp2}.

Tab.~\ref{tab1} shows that our results achieve better PSNR and SSIM than the state-of-the-art methods on seen styles and unseen styles. 
This performance also coincides with the visual results. 

\noindent{\textbf{Consistency Analysis. }}
Intuitively, the destylized faces from the different styles of the same person should look similar. Examples generated from multiple styles are shown in Fig.~\ref{fig:cmp1g} and Fig.~\ref{fig:cmp2g}. In this experiment, we demonstrate that our method not only recovers realistic faces with high fidelity but also generates faces looking close to each other given multiple styles of the same person on input. This indicates that SRN can indeed extract facial features from portraits despite different styles and transfer these features to recover underlying faces. 

\begin{table}\renewcommand{\arraystretch}{1.23}
\caption{Comparison of consistency between destylized faces from various seen and unseen styles.}
\begin{center}
\begin{tabular}{|c|c|c|}
\hline
& Seen Styles & Unseen Styles\\
\hline
Gatys ~\cite{gatys2016image}  & 82\%  & 83\% \\ 
Johnson~\cite{johnson2016perceptual} & 73\%  &  72.5\%  \\
MGAN~\cite{li2016precomputed}  & 2\%  &  1\% \\
pix2pix~\cite{isola2016image} & 93.33\%  &  85.1\%  \\
\hline
Ours & {\bf 98\%} & {\bf 90.8\%} \\
\hline
\end{tabular}
\label{tab2}
\end{center}
\vspace{-0.3cm}
\end{table}
To evaluate the consistency of generated faces from different portraits of the same person, we adapt the off-the-shelf deep face recognition approach~\cite{parkhi2015deep}. 
First, we randomly choose 100 RF and 800 corresponding SF faces from eight different styles in the test dataset for our gallery (three seen styles and five unseen styles). Then, we employ Gatys ~\cite{gatys2016image}, Johnson~\cite{johnson2016perceptual}, MGAN~\cite{li2016precomputed}, pix2pix~\cite{isola2016image} and our FDNN to recover real faces from eight various stylized faces. For each method, we set 100 destylized faces from the \emph{Candy} style as a query dataset and set the other 700 destylized faces from the other seven styles as a search dataset. Following the standard protocol, we compute the Face Recognition Rate (FRR) which quantifies if the correct person is retrieved within the top-5 candidates (the probability of successful retrieval by chance is 0.71\%). 
We also use the same procedure for other styles. Table~\ref{tab2} shows the average FRR of each method for seen and unseen styles. Our method yields high consistency score for both seen and unseen styles. This indicates the effectiveness of our FDNN in producing realistic faces of high-fidelity.


\begin{figure}
\centering
\includegraphics[width=0.91\linewidth]{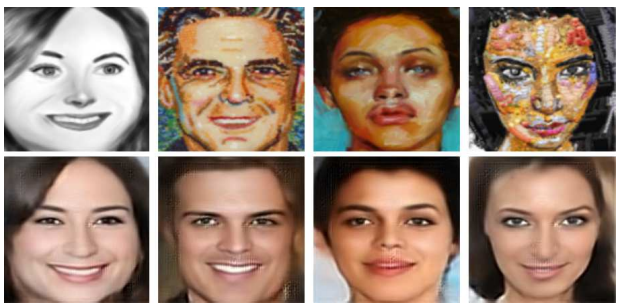}
\caption{Results for the original paintings. Top row: the original portraits from DevianArt. Bottom row: our destylization results.}
\label{fig:Orig}
\end{figure}

\subsection{Performance on Original Paintings}
Despite our method is trained on a synthetic dataset, it can efficiently generalize to real paintings/portraits. To demostrate this, we randomly choose some paintings with faces from DevianArt. We crop images of these faces and then align them to the CelebA face dataset in an off-line pre-processing step.
Our method successfully reconstructs plausible facial details from real paintings as shown in Fig.~\ref{fig:Orig}. This highlights that our method is not restricted to synthesized stylized faces. 

\begin{figure}
\vspace{-0.2cm}
\centering
\subfigure[Unaligned]{\label{fig:Faila}\scalebox{1}[1]{\includegraphics[width=0.24\linewidth]{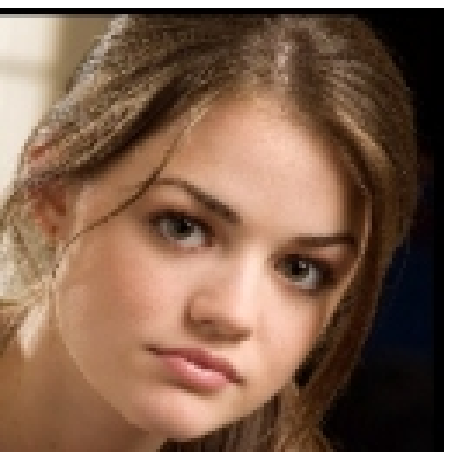}}}
\subfigure[SF]{\label{fig:Failb}\scalebox{1}[1]{\includegraphics[width=0.24\linewidth]{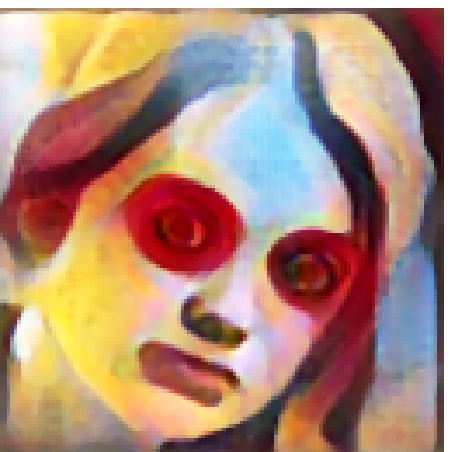}}}
\subfigure[Our result]{\label{fig:Failc}\scalebox{1}[1]{\includegraphics[width=0.24\linewidth]{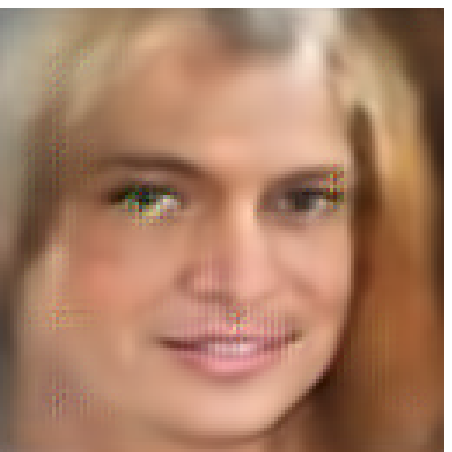}}}\\\vspace{-0.5em}
\subfigure[Upright pose]{\label{fig:Faild}\scalebox{1}[1]{\includegraphics[width=0.24\linewidth]{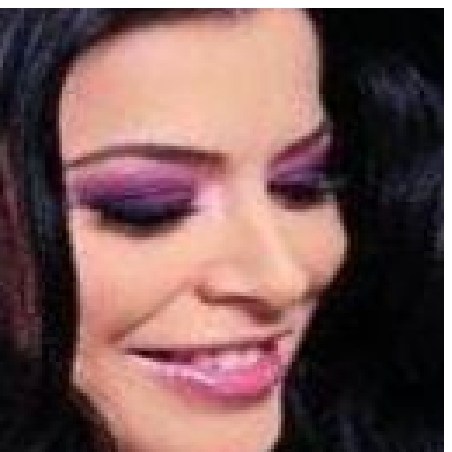}}}
\subfigure[SF]{\label{fig:Faile}\scalebox{1}[1]{\includegraphics[width=0.24\linewidth]{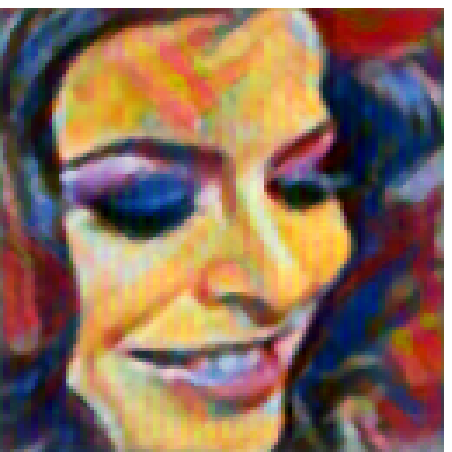}}}
\subfigure[Our result]{\label{fig:Failf}\scalebox{1}[1]{\includegraphics[width=0.24\linewidth]{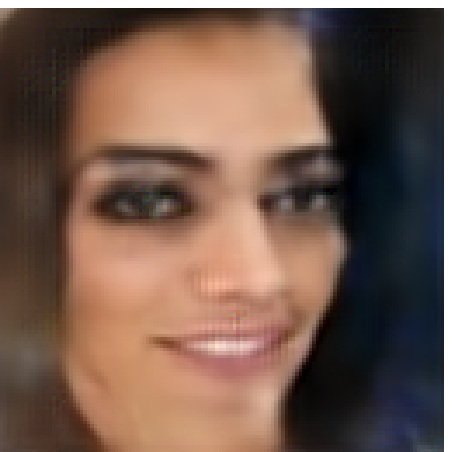}}}
\caption{Failures. (a) An unaligned ground-truth face. (e) Stylized face of (a). (c) Our result. (d) An upright pose. (e) Stylized face of (d). (c) Our result.}
\label{fig:Failure}
\vspace{-0.3cm}
\end{figure}

\subsection{Limitations} 

Our proposed network requires that the eyes of stylized faces to be aligned beforehand to a template. Without such an alignment, FDNN may generate artifacts. However, we plan to automatically align the stylized facial images in our future work. As illustrated in Fig.~\ref{fig:Faila}, destylization is performed on an unaligned stylized face. As a consequence, our network cannot localize facial features correctly and produces erroneous feature maps. 
In addition, our method may produce artifacts for portraits suffering from large pose variations, such as profile views of faces etc. Since there are not enough side-view images of faces in the training dataset, this results in artifacts. As shown in Fig.~\ref{fig:Failf}, the network fails to generate satisfying results for an upright pose. Exploring how to address large pose variations will be our future work.

\section{Conclusion}
We presented a face destylization method that extracts features of a stylized portrait and then exploits them to generate its corresponding photo-realistic face. Our network learns a mapping from stylized facial feature maps to realistic facial feature maps. Our network can successfully extract facial features from different styles and thus is able to destylize unseen style portraits as well.

\section*{Acknowledgement}
This work is supported by the Australian Research Council (ARC) grant DP150104645.

\ifCLASSOPTIONcaptionsoff
  \newpage
\fi

\bibliographystyle{IEEEtran}
\bibliography{IEEEabrv}

\end{document}